\documentclass{article}

\usepackage[nonatbib,final]{neurips_2022}

\usepackage[utf8]{inputenc} 
\usepackage[T1]{fontenc}    
\usepackage{hyperref}       
\usepackage{url}            
\usepackage{booktabs}       
\usepackage{amsfonts}       
\usepackage{nicefrac}       
\usepackage{microtype}      
\usepackage{hyperref}

\usepackage[ruled,vlined,linesnumbered]{algorithm2e}
\usepackage{xspace}
\usepackage{booktabs}
\usepackage{colortbl}
\usepackage{mathtools}
\usepackage{amsmath}
\usepackage{amssymb}
\usepackage{amsthm}
\usepackage{amsfonts}
\usepackage{cleveref}
\usepackage[fleqn]{nccmath}
\Crefformat{section}{Section~#2#1#3}
\Crefformat{proposition}{Proposition~#2#1#3}
\Crefformat{equation}{Eq.\,#2#1#3}

\usepackage{bm}
\usepackage{xspace}
\usepackage{enumitem}
\usepackage[dvipsnames]{xcolor}
\usepackage{subcaption}
\usepackage{caption}
\usepackage{wrapfig}
\usepackage{stackengine}
\usepackage{multirow}
\usepackage{array}
\usepackage{pbox}
\usepackage{dcolumn}
\usepackage[final]{pdfpages}

\usepackage{dsfont}
\usepackage{pifont}
\usepackage{marvosym}
\usepackage{makecell}
\usepackage{cellspace}
\usepackage[super]{nth}

\usepackage{listings}

\newcommand{\xmark}{\ding{55}}%
\newcolumntype{Y}{>{\centering\arraybackslash}X}

\newlength\savewidth\newcommand\shline{\noalign{\global\savewidth\arrayrulewidth
  \global\arrayrulewidth 1pt}\hline\noalign{\global\arrayrulewidth\savewidth}}

\makeatletter\renewcommand\paragraph{\@startsection{paragraph}{4}{\z@}
  {.5em \@plus1ex \@minus.2ex}{-.5em}{\normalfont\normalsize\bfseries}}\makeatother
\definecolor{pearDark}{HTML}{2980B9}
\definecolor{pearDarker}{HTML}{1D2DEC}

\definecolor{codegreen}{rgb}{0,0.6,0}
\definecolor{codegray}{rgb}{0.5,0.5,0.5}
\definecolor{codepurple}{rgb}{0.58,0,0.82}
\definecolor{backcolour}{rgb}{0.95,0.95,0.92}

\lstdefinestyle{mystyle}{
    backgroundcolor=\color{backcolour},   
    commentstyle=\color{codegreen},
    keywordstyle=\color{magenta},
    numberstyle=\tiny\color{codegray},
    stringstyle=\color{codepurple},
    basicstyle=\linespread{1}\ttfamily\tiny,
    breakatwhitespace=false,
    breaklines=true,                 
    captionpos=b,                    
    keepspaces=true,                 
    numbers=left,                    
    numbersep=1pt,                  
    showspaces=false,                
    showstringspaces=false,
    showtabs=false,
    tabsize=2
}

\newcommand{\minus}{\scalebox{0.75}[1.0]{$-$}}

\newcommand{\removed}[1]{}

\def\eqref#1{equation~\ref{#1}}

\DeclareMathAlphabet{\mathsfit}{\encodingdefault}{\sfdefault}{m}{sl}
\SetMathAlphabet{\mathsfit}{bold}{\encodingdefault}{\sfdefault}{bx}{n}

\newcommand{\sqboxs}{1.2ex}
\newcommand{\sqbox}[1]{\textcolor{#1}{\rule{\sqboxs}{\sqboxs}}}
\newcommand{\sqboxblack}[1]{\setlength{\fboxsep}{0pt}\fbox{\sqbox{#1}}}

\newcommand{\blue}[1]{{\textcolor{black}{#1}}}

\newcommand{\ie}{{\emph{i.e.}}\xspace}
\newcommand{\eg}{{\emph{e.g.}}\xspace}

\newcommand\paperurl[1]{{\scriptsize{\color{blue}{\url{#1}}}}}

\title{Expediting Large-Scale Vision Transformer\\ for Dense Prediction without Fine-tuning}

\author{
Weicong Liang\textsuperscript{1}\thanks{
Equal contribution.}
\enskip \enskip \quad
Yuhui Yuan\textsuperscript{4}$^*
\thanks{\Letter\; {yuhui.yuan}@microsoft.com}$
\enskip  \enskip \quad
\textbf{Henghui Ding\textsuperscript{3}}
\enskip \enskip \quad
Xiao Luo\textsuperscript{2} \\
\textbf{Weihong Lin\textsuperscript{4}}
\enskip
\textbf{Ding Jia\textsuperscript{1}}
\enskip \enskip 
\textbf{Zheng Zhang\textsuperscript{4}}
\enskip \enskip 
\textbf{Chao Zhang\textsuperscript{1}}
\enskip \enskip
\textbf{Han Hu\textsuperscript{4}}
\\
\textsuperscript{1}Key Laboratory of Machine Perception (MOE) \\
School\: of\: Intelligence\: Science\: and\: Technology,\; Peking University \\
\textsuperscript{2}School of Mathematical Sciences, Peking University
\enskip\space
\textsuperscript{3}ETH Zurich\\
\textsuperscript{4}Microsoft Research Asia
\enskip\enskip\enskip\quad
}

\begin{document}

\maketitle

\begin{abstract}
Vision transformers have recently achieved competitive results across various vision tasks but still suffer from heavy computation costs when processing a large number of tokens. Many advanced approaches have been developed to reduce the total number of tokens in large-scale vision transformers, especially for image classification tasks. Typically, they select a small group of essential tokens according to their relevance with the [\texttt{class}] token, then fine-tune the weights of the vision transformer. Such fine-tuning is less practical for dense prediction due to the much heavier computation and GPU memory cost than image classification.

In this paper, we focus on a more challenging problem, \ie, accelerating large-scale vision transformers for dense prediction without any additional re-training or fine-tuning. In response to the fact that high-resolution representations are necessary for dense prediction, we present two non-parametric operators, a \emph{token clustering layer} to decrease the number of tokens and a \emph{token reconstruction layer} to increase the number of tokens. The following steps are performed to achieve this: (i) we use the token clustering layer to cluster the neighboring tokens together, resulting in low-resolution representations that maintain the spatial structures; (ii) we apply the following transformer layers only to these low-resolution representations or clustered tokens; and (iii) we use the token reconstruction layer to re-create the high-resolution representations from the refined low-resolution representations. The results obtained by our method are promising on five dense prediction tasks, including object detection, semantic segmentation, panoptic segmentation, instance segmentation, and depth estimation. Accordingly, our method accelerates $40\%\uparrow$ FPS and saves $30\%\downarrow$ GFLOPs of ``Segmenter+ViT-L/$16$'' while maintaining $99.5\%$ of the performance on ADE$20$K without fine-tuning the official weights.
\end{abstract}

\section{Introduction}
\label{intro}

Transformer~\cite{vaswani2017attention} has made significant progress across various challenging vision tasks since pioneering efforts such as DETR~\cite{carion20}, Vision Transformer (ViT)~\cite{dosovitskiy2020image}, and Swin Transformer~\cite{liu2021swinv1}.
By removing the local inductive bias~\cite{d2021convit} from convolutional neural networks~\cite{He2016ResNet,szegedy2015going,simonyan2014very}, vision transformers armed with global self-attention show superiority in scalability for large-scale models and billion-scale dataset~\cite{dosovitskiy2020image,zhai2022lscalevit,singh2022revisiting}, self-supervised learning~\cite{he2021masked,xie2021simmim,bao2021beit}, connecting vision and language~\cite{radford2021learning,jia2021scaling}, etc.
We can find from recent developments of SOTA approaches that vision transformers have dominated various leader-boards, including but not limited to
image classification~\cite{wortsman2022model,dai2021coatnet,ding2022davit,zhai2022lscalevit}, object detection~\cite{zhang2022dino,liu2021swinv2,li2021grounded}, semantic segmentation~\cite{jain2021semask,cheng2021mask2former,bousselham2021efficient}, pose estimation~\cite{xu2022vitpose}, image generation~\cite{zhang2021styleswin}, and depth estimation~\cite{li2022depthformer}.

Although vision transformers have achieved more accurate predictions in many vision tasks, large-scale vision transformers are still burdened with heavy computational overhead, particularly when processing high-resolution inputs~\cite{el2021xcit,liu2021swinv2}, thus limiting their broader application to more resource-constrained applications and attracting efforts on re-designing light-weight vision transformer architectures~\cite{chen2021mobile,mehta2021mobilevit,zhang2022edgeformer}.
In addition to this, several recent efforts have investigated how to decrease the model complexity and accelerate vision transformers, especially for image classification, and introduced various advanced approaches to accelerate vision transformers.
Dynamic ViT~\cite{rao2021dynamicvit} and EViT~\cite{liang2021evit}, for example, propose two different dynamic token sparsification frameworks to reduce the redundant tokens progressively and select the most informative tokens according to the scores predicted with an extra trained prediction module or their relevance with the [\texttt{class}] token.
TokenLearner~\cite{ryoo2021tokenlearner} learns to spatially attend over a subset of tokens and generates a set of clustered tokens adaptive to the input for video understanding tasks.
Most of these token reduction approaches are carefully designed for image classification tasks and 
require fine-tuning or retraining. These approaches might not be suitable to tackle more challenging dense prediction tasks that need to process high-resolution input images, e.g., $1024\times1024$, thus, resulting in heavy computation and GPU memory cost brought.
We also demonstrate in the supplemental material the superiority of our method over several representative methods on dense prediction tasks.

\begin{figure*}[t]
\begin{minipage}[t]{1\linewidth}
\centering
\hspace{-5mm}
\begin{subfigure}[b]{0.33\textwidth}
{\includegraphics[width=\textwidth,height=25mm]{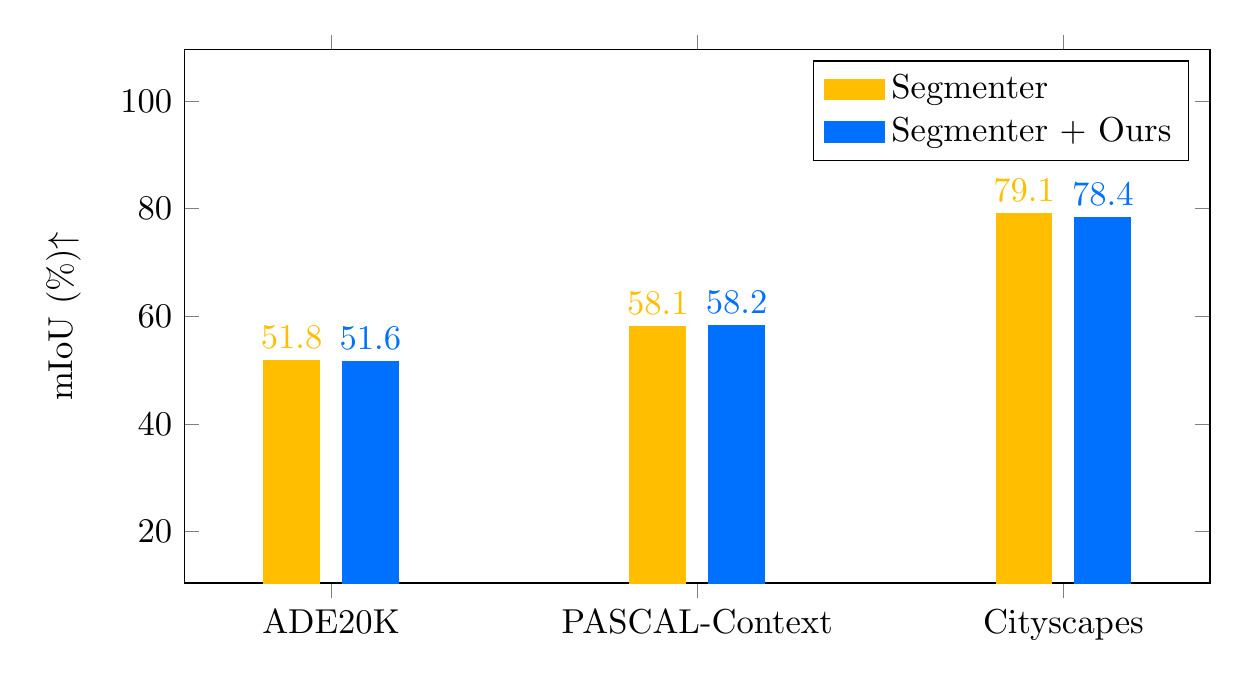}}
\caption{\small{mIoU@Segmenter}}
\end{subfigure}
\begin{subfigure}[b]{0.33\textwidth}
{\includegraphics[width=\textwidth,height=25mm]{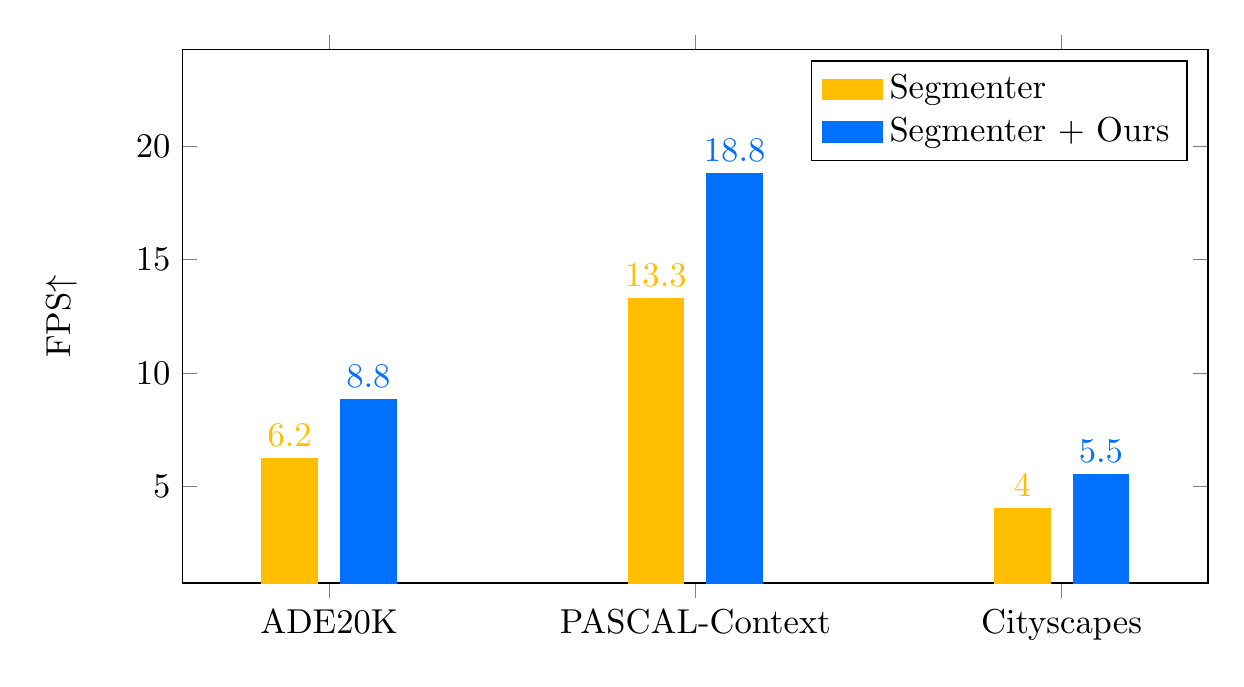}}
\caption{\small{FPS@Segmenter}}
\end{subfigure}
\begin{subfigure}[b]{0.33\textwidth}
{\includegraphics[width=\textwidth,height=25mm]{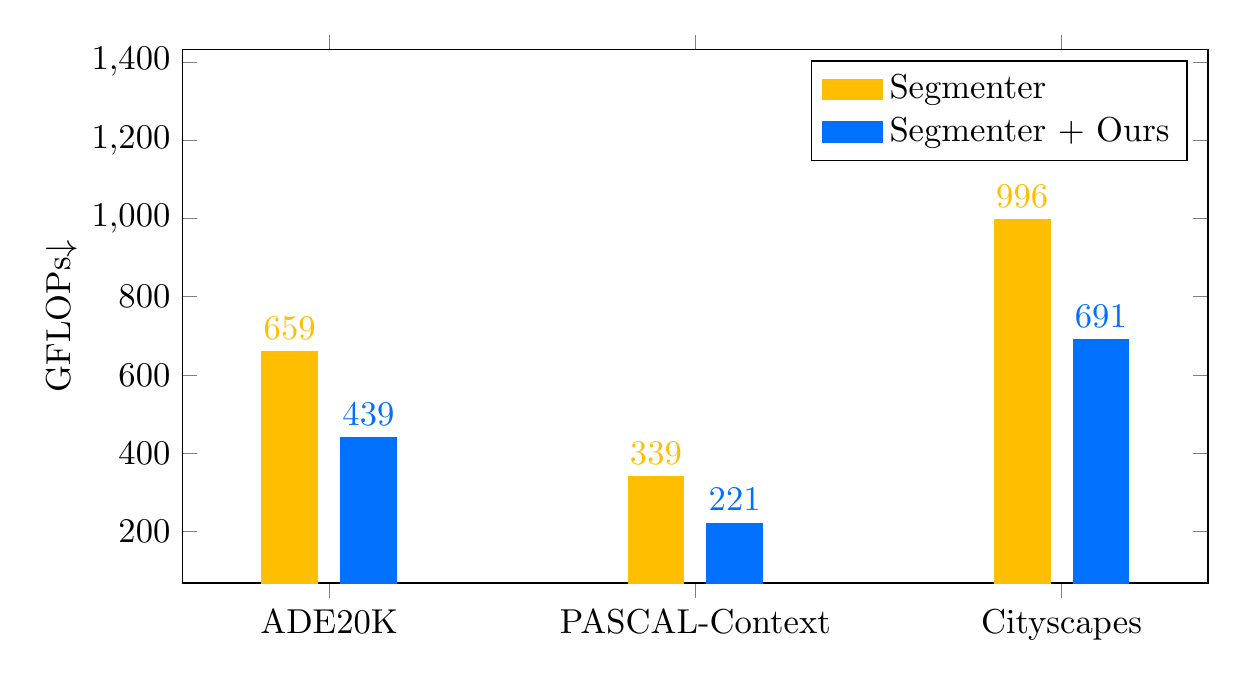}}
\caption{\small{GFLOPs@Segmenter}}
\end{subfigure}\\
\hspace{-5mm}
\begin{subfigure}[b]{0.33\textwidth}
{\includegraphics[width=\textwidth,height=25mm]{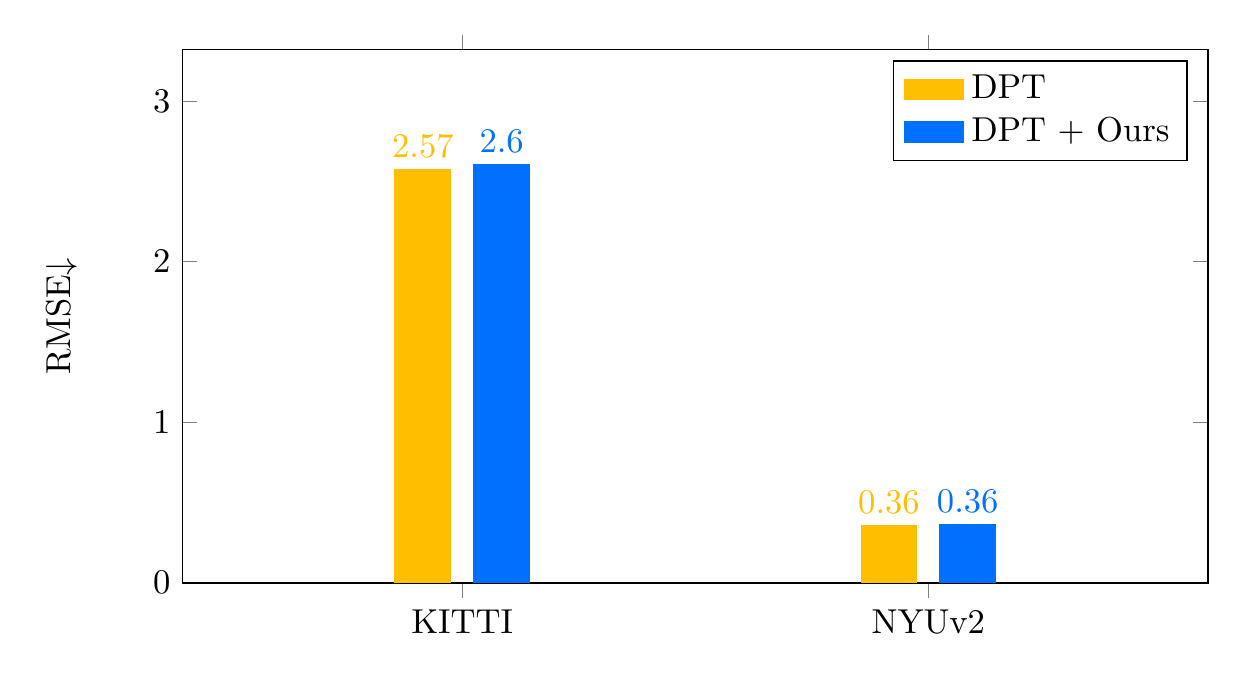}}
\caption{\small{RMSE@DPT}}
\end{subfigure}
\begin{subfigure}[b]{0.33\textwidth}
{\includegraphics[width=\textwidth,height=25mm]{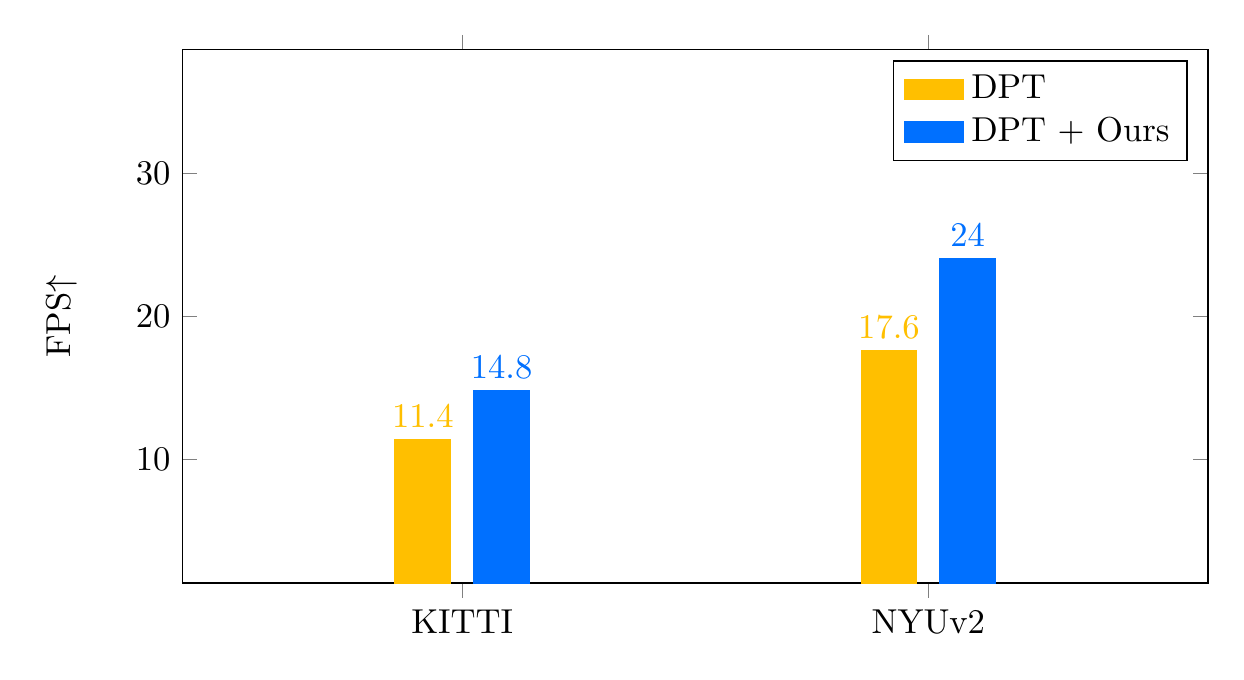}}
\caption{\small{FPS@DPT}}
\end{subfigure}
\begin{subfigure}[b]{0.33\textwidth}
{\includegraphics[width=\textwidth,height=25mm]{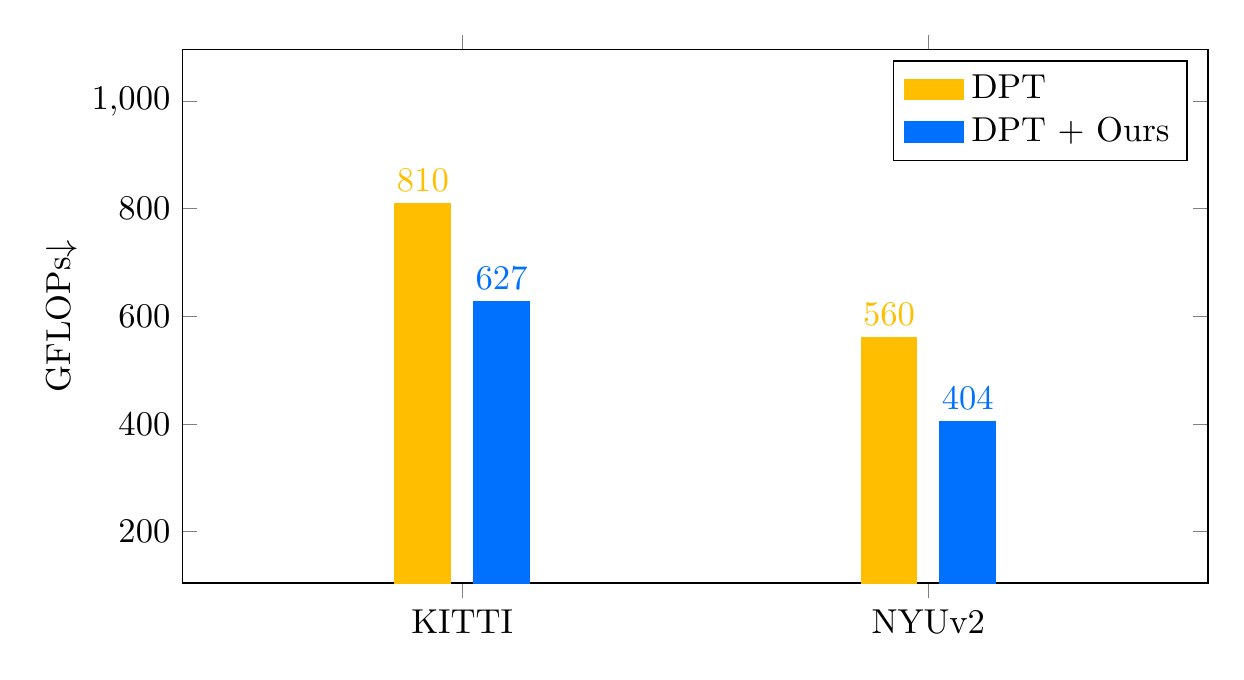}}
\caption{\small{GFLOPs@DPT}}
\end{subfigure}
\caption{Illustrating the improvements of our approach: we report the results of applying our approach to Segmenter~\cite{Segmenter} for semantic segmentation and DPT~\cite{ranftl2021vision} for depth estimation on the $1$-st and $2$-ed row respectively. Without any fine-tuning, our proposed method reduces the GFLOPs and accelerates the FPS significantly with a slight performance drop on both dense prediction tasks. $\uparrow$ and $\downarrow$ represent higher is better and lower is better respectively. Refer to Section~\ref{sec:experiment} for more details.}
\label{fig:intro_results}
\end{minipage}
\vspace{-3mm}
\end{figure*}

Rather than proposing a new lightweight architecture for dense prediction or token reduction scheme for only image classification, we focus on how to expedite well-trained large-scale vision transformers and use them for various dense prediction tasks without fine-tuning or re-training.
Motivated by these two key observations including (i) the intermediate token representations of a well-trained vision transformer carry a heavy amount of local spatial redundancy and (ii) dense prediction tasks require high-resolution representations,
we propose a simple yet effective scheme to convert the ``high-resolution'' path of the vision transformer to a ``high-to-low-to-high resolution'' path via two non-parametric layers including a token clustering layer and a token reconstruction layer.
Our method can produce a wide range of more efficient models without requiring further fine-tuning or re-training.
We apply our approach to expedite two main-stream vision transformer architectures, e.g., ViTs and Swin Transformers, for five challenging dense prediction tasks, including object detection, semantic segmentation, panoptic segmentation, instance segmentation, and depth estimation.
We have achieved encouraging results across several evaluated benchmarks
and Figure~\ref{fig:intro_results} illustrates some representative results on both semantic segmentation and depth estimation tasks.

\section{Related work}
\label{relatework}

\paragraph{Pruning Convolutional Neural Networks.} Convolutional neural network pruning~\cite{blalock2020state,hoefler2021sparsity,wang2021accelerate} is a task that involves removing the redundant parameters to reduce the model complexity without a significant performance drop.
Pruning methods typically entail three steps: (i) training a large, over-parameterized model to convergence, (ii) pruning the trained large model according to a certain criterion, and (iii) fine-tuning the pruned model to regain the lost performance~\cite{liu2018rethinking}.
The key idea is to design an importance score function that is capable of pruning the less informative parameters.
We follow~\cite{chen2021chasing} to categorize the existing methods into two main paths: (i) unstructured pruning (also named weight pruning) and (ii) structured pruning. Unstructured pruning methods explore the absolute value of each weight or the product of each weight and its gradient to estimate the importance scores. Structured pruning methods, such as layer-level pruning~\cite{wang2019dbp}, filter-level pruning\cite{liu2017learning,yu2018slimmable}, and image-level pruning~\cite{han2020model,tan2019efficientnet}, removes the model sub-structures. Recent studies~\cite{chavan2022vision,chen2021autoformer,hou2021multi} further extend these pruning methods to vision transformer.
Unlike the previous pruning methods, we explore how to expedite vision transformers for dense prediction tasks by carefully reducing \& increasing the number of tokens without removing or modifying the parameters.

\paragraph{Efficient Vision Transformer.}
The success of vision transformers has incentivised many recent efforts~\cite{ryoo2021tokenlearner,song2021dynamic,marin2021token,heo2021rethinking,wang2021not,rao2021dynamicvit,kong2021spvit,guibas2021efficient,wang2021pnp,chen2021auto,kahatapitiya2021swat,liang2021evit,roy2021efficient} to exploit the spatial redundancies of intermediate token representations.
For example,
TokenLearner~\cite{ryoo2021tokenlearner} learns to attend over a subset of tokens and generates a set of clustered tokens adaptive to the input.
They empirically show that very few clustered tokens are sufficient for video understanding tasks.
Token Pooling~\cite{marin2021token} exploits a nonuniform data-aware down-sampling operator based on K-Means or K-medoids to cluster similar tokens together to reduce the number of tokens while minimizing the reconstruction error.
Dynamic ViT~\cite{rao2021dynamicvit} observes that the accurate image recognition with vision transformers mainly depends on a subset of the most informative tokens,
and hence it develops a dynamic token sparsification framework for pruning the redundant tokens dynamically based on the input.
EViT (expediting vision transformers)~\cite{liang2021evit} proposes to calculate
the attentiveness of the [\texttt{class}] token with respect to each token and identify the top-$k$ attentive tokens according to the attentiveness score.
Patch Merger~\cite{renggli2022learning} uses a learnable attention matrix to merge and combine together the redundant tokens, therefore creating a much more practical and cheaper model with only a slight performance drop.
Refer to~\cite{tay2020efficient} for more details on efficient transformer architecture designs,
such as Performer~\cite{choromanski2020rethinking} and Reformer~\cite{kitaev2020reformer}.
In contrast to these methods that require either retraining or fine-tuning the modified transformer architectures from scratch or the pre-trained weights, our approach can reuse the once-trained weights for free and produce lightweight models with a modest performance drop.

\paragraph{Vision Transformer for Dense Prediction.}
In the wake of success of the representative pyramid vision transformers~\cite{liu2021swinv1,wang2021pyramid} for object detection and semantic segmentation, more and more efforts have explored different advanced vision transformer architecture designs~\cite{li2022exploring,chen2021simple,li2021improved,li2021benchmarking,fan2021multiscale,yuan2021hrformer,yang2021focal,zhang2021multi,gu2021hrvit,liang2021swinir,xie2021segformer,yuan2020object,dingVLT,yuan2020segfix,yuan2018ocnet,he2022mlseg} suitable for various dense prediction tasks.
For example, MViT~\cite{li2021improved} focuses more on multi-scale representation learning, while HRFormer~\cite{yuan2021hrformer} examines the benefits of combining multi-scale representation learning and high-resolution representation learning.
Instead of designing a novel vision transformer architecture for dense prediction,
we focus on how to accelerate a well-trained vision transformer while maintaining the prediction performance as much as possible.

\paragraph{Our approach.}
The contribution of our work lies in two main aspects:
(i) we are the first to study how to accelerate state-of-the-art large-scale vision transformers for dense prediction tasks without fine-tuning (e.g., "Mask2Former + Swin-L" and "SwinV2-L + HTC++"). Besides, our approach also achieves much better accuracy and speedup trade-off when compared to the very recent ACT [1] which is based on a clustering attention scheme;
(ii) our token clustering and reconstruction layers are capable of maintaining the semantic information encoded in the original high-resolution representations. This is the very most important factor to avoid fine-tuning.
We design an effective combination of a token clustering function and a token reconstruction function to maximize the cosine similarity between the reconstructed high-resolution feature maps and the original ones without fine-tuning.
The design of our token reconstruction layer is the key and not straightforward essentially. We also show that our token reconstruction layer can be used to adapt the very recent EViT~\cite{liang2021evit} and DynamicViT~\cite{rao2021dynamicvit} for dense prediction tasks in the supplementary.

\section{Our Approach}
\label{approach}

\begin{figure*}[t]
\definecolor{myyellow}{RGB}{255, 204, 153}
\definecolor{myblue}{RGB}{204,229,255}
\hspace{5mm}
\begin{subfigure}[b]{0.39\linewidth}
\centering
\includegraphics[width=0.99\linewidth]{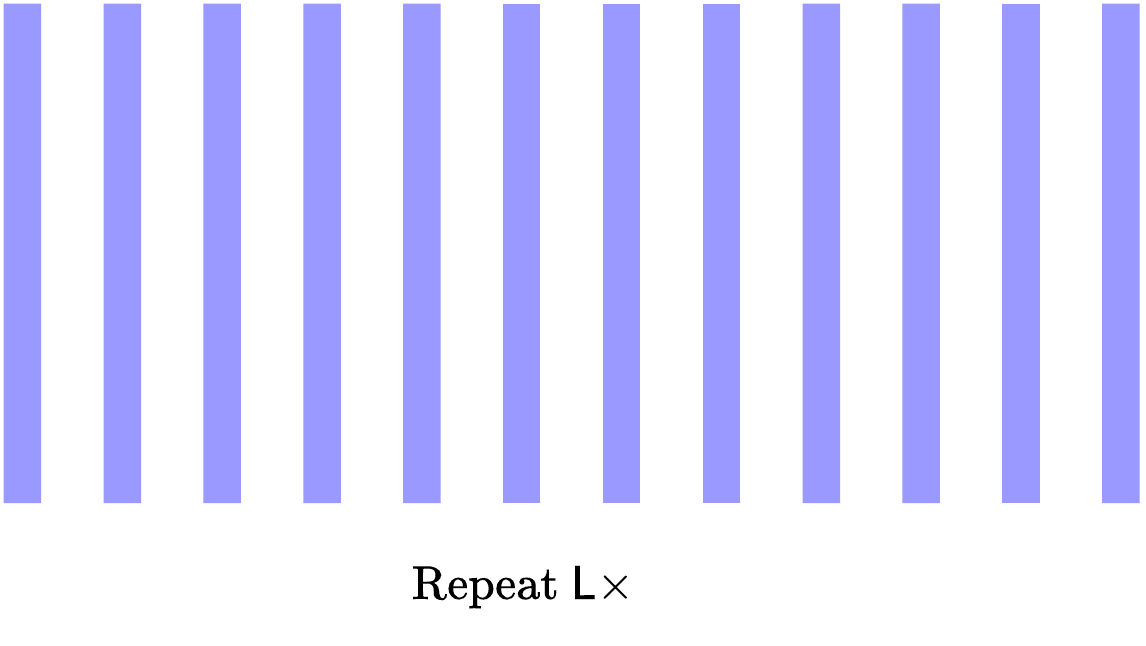}
\caption{\small{}}
\end{subfigure}
\begin{subfigure}[b]{0.6\linewidth}
\centering
\hspace{5mm}
\includegraphics[width=0.98\linewidth]{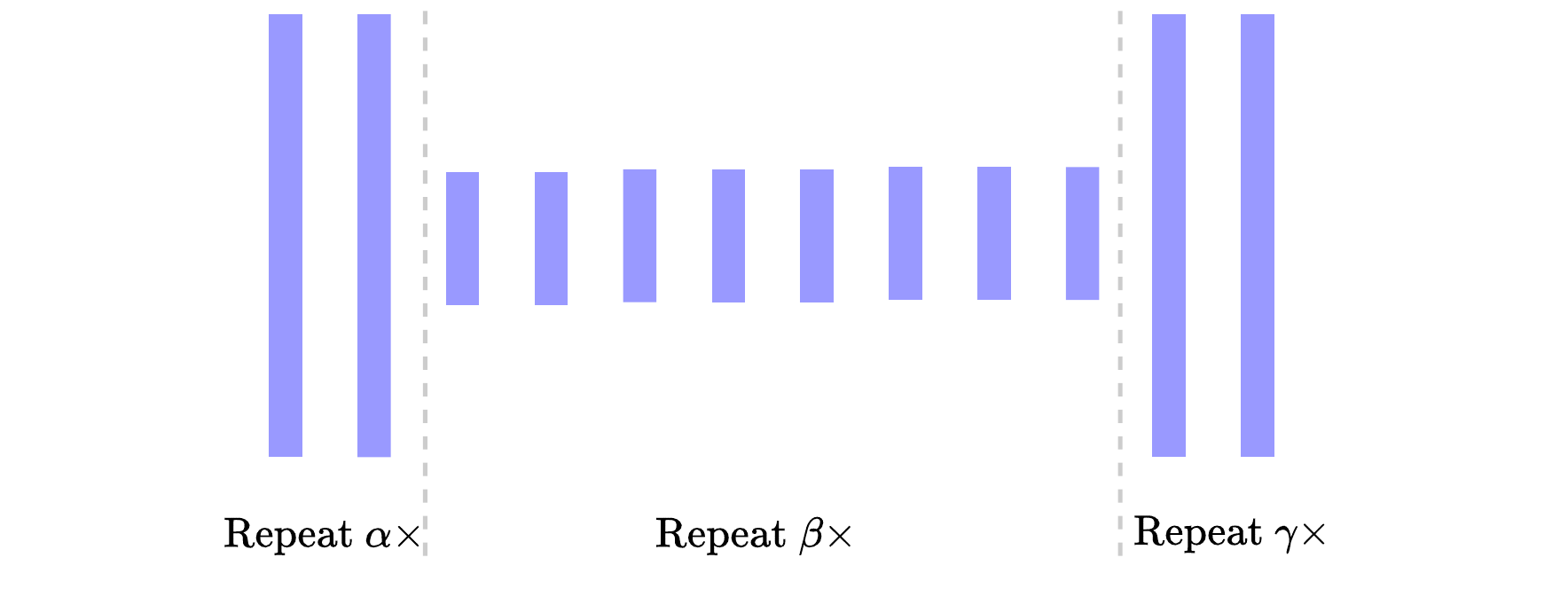}
\caption{}
\end{subfigure}
\begin{subfigure}[b]{1\textwidth}
\hspace{-2mm}
\includegraphics[width=1\textwidth]{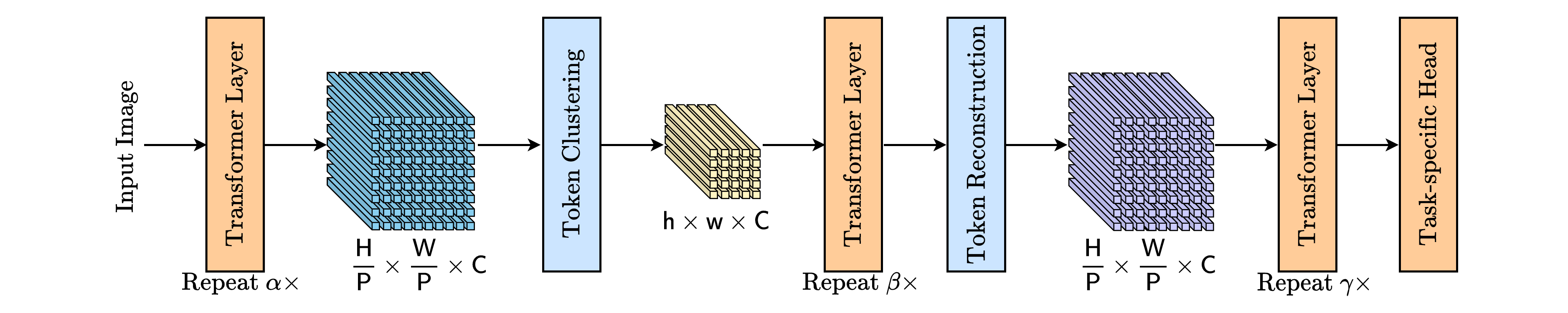}
\caption{}
\end{subfigure}
\caption{(a) Plain high-resolution vision transformer with $\mathsf{L}$ layers. (b) U-shape High-to-low-to-high-resolution vision transformer with $\alpha$, $\beta$, and $\gamma$ layers respectively ($\mathsf{L}=\alpha+\beta+\gamma$). (c) Illustrating the details of using our approach to plain ViTs:
we insert a token clustering layer and a token reconstruction layer into a trained vision transformer in order to decrease and then increase the spatial resolution, respectively.
The weights of modules marked with \sqboxblack{myyellow}
are trained once based on the configuration of (a).
The token clustering layer and token reconstruction layer are marked with \sqboxblack{myblue}
are non-parametric,
thus do not require any fine-tuning and can be included directly during evaluation.
}
\label{fig:freeformer}
\end{figure*}

\noindent\textbf{Preliminary.}
The conventional Vision Transformer~\cite{dosovitskiy2020image} first reshapes the input image $\mathbf{X} \in \mathbb{R}^{\mathsf{H}\times \mathsf{W}\times \mathsf{3}}$ into a sequence of flatten patches $\mathbf{X}_p \in \mathbb{R}^{\mathsf{N}\times (\mathsf{P}^2 \mathsf{3})}$,
where $(\mathsf{P}, \mathsf{P})$ represents the resolution of each patch,
$(\mathsf{H}, \mathsf{W})$ represents the resolution of the input image,
$\mathsf{N}=(\mathsf{H}\times \mathsf{W})/{\mathsf{P}^2}$ represents the number of resulting patches or tokens, \ie, the input sequence length.
The Vision Transformer consists of alternating layers of multi-head self-attention ($\operatorname{MHSA}$)
and feed-forward network ($\operatorname{FFN}$) accompanied with layer norm ($\operatorname{LN}$) and residual connections:
\begin{equation}
\begin{aligned} {\mathbf{Z}_{l}^{'}}&=\operatorname{MHSA}\left(\mathrm{L N}\left(\mathbf{Z}_{l-1}\right)\right)+\mathbf{Z}_{l-1}, \\ {\mathbf{Z}_{l}} &=\operatorname{FFN}\left(\mathrm{LN}\left({\mathbf{Z}_{l}^{'}}\right)\right)+{\mathbf{Z}_{l}^{'}}, 
\end{aligned}
\end{equation}
where $l \in \{1,\dots, \mathsf{L}\}$ represents the layer index,
$\mathbf{Z}_{l}\in \mathbb{R}^{\mathsf{N}\times\mathsf{C}}$, and $\mathbf{Z}_{0}$ is based on $\mathbf{X}_p$. The computation cost $\mathcal{O}(\mathsf{L}\mathsf{N}\mathsf{C}(\mathsf{N}+\mathsf{C}))$ mainly depends on the number of layers $\mathsf{L}$, the number of tokens $\mathsf{N}$,
and the channel dimension $\mathsf{C}$.

Despite the great success of transformer, its computation cost increases significantly when handling high-resolution representations, which are critical for dense prediction tasks.
This paper attempts to resolve this issue by reducing the computation complexity during the inference stage,
and presents a very simple solution for generating a large number of efficient vision transformer models directly from a single trained vision transformer, requiring no further training or fine-tuning.

We demonstrate how our approach could be applied to the existing standard Vision Transformer in \figurename~\ref{fig:freeformer}. The original Vision Transformer is modified using two non-parametric operations, namely a token clustering layer and a token reconstruction layer.
The proposed token clustering layer is utilized to convert the high-resolution representations to low-resolution representations by clustering the locally semantically similar tokens. Then, we apply the following transformer layers on the low-resolution representations, which greatly accelerates the inference speed and saves computation resources. Last, a token reconstruction layer is proposed to reconstruct the feature representations back to high-resolution.

\vspace{2mm}
\noindent\textbf{Token Clustering Layer.}
We construct the token clustering layer following the improved SLIC scheme~\cite{jampani2018superpixel}, which performs local k-means clustering as follows:

\vspace{1mm}
\noindent-\emph{Initial superpixel center}:
We apply adaptive average pooling ($\operatorname{AAP}$) over the high-resolution representations from the $\alpha$-th layer to compute
the $\mathsf{h}\times \mathsf{w}$ initial cluster center representations:
\begin{equation}
\mathbf{S}_{\alpha} = \operatorname{AAP}(\mathbf{Z}_{\alpha}, (\mathsf{h}\times \mathsf{w})),\label{eq:pooling}
\end{equation}
where $\mathbf{S}_{\alpha}\in \mathbb{R}^{\mathsf{h w}\times\mathsf{C}}$,
$\mathbf{Z}_{\alpha}\in \mathbb{R}^{\mathsf{N}\times\mathsf{C}}$, and $\mathsf{h w}\ll \mathsf{N}$.

\vspace{1mm}
\noindent-\emph{Iterative local clustering}:
(i) Expectation step: compute the normalized similarity between each pixel $p$ and the surrounding superpixel $i$ (we only consider the neighboring $\lambda$ positions),
(ii) Maximization step: compute the new superpixel centers:
\begin{equation}
\begin{aligned}
{\mathbf{Q}_{p,i}} = \frac{\rm{exp}\Big(\minus ||\mathbf{Z}_{\alpha,p} - \mathbf{S}_{\alpha,i}||^2 /{\tau} \Big)}{\sum_{j=1}^{\lambda} \rm{exp}\Big(\minus||\mathbf{Z}_{\alpha,p} - \mathbf{S}_{\alpha,j}||^2 /{\tau}\Big)},\quad
\mathbf{S}_{\alpha,i} = \sum_{p=1}^\mathsf{N} \mathbf{Q}_{p,i} \mathbf{Z}_{\alpha,p},\label{eq:token_cluster}
\end{aligned}
\end{equation}
where we iterate the above Expectation step and Maximization step for $\kappa$ times, $\tau$ is a temperature hyper-parameter, and $i\in\{1,2,\cdots,\lambda\}$.
We apply the following $\beta$ transformer layers on $\mathbf{S}_{\alpha}$ instead of $\mathbf{Z}_{\alpha}$,
thus results in $\mathbf{S}_{\alpha+\beta}$ and decreases the computation cost significantly.

\vspace{2mm}
\noindent\textbf{Token Reconstruction Layer.}
We implement the token reconstruction layer by exploiting the relations between the high-resolution representations
and the low-resolution clustered representations:
\begin{equation}
\begin{aligned}
\mathbf{Z}_{\alpha+\beta,p} = \sum_{\mathbf{S}_{\alpha,i} \in \textrm{k-NN}(\mathbf{Z}_{\alpha,p})} \frac{\textrm{exp}\big(\minus{||\mathbf{Z}_{\alpha,p} - \mathbf{S}_{\alpha,i}||^2}/{\tau}\big)}{\sum_{\mathbf{S}_{\alpha,j} \in \textrm{k-NN}(\mathbf{Z}_{\alpha,p})}\textrm{exp}\big(\minus||\mathbf{Z}_{\alpha,p} - \mathbf{S}_{\alpha,j}||^2/{\tau}\big)} \mathbf{S}_{\alpha+\beta,i},
\end{aligned}\label{eq:token_reconstruct}
\end{equation}
where $\tau$ is the same temperature hyper-parameter as in Equation~\ref{eq:token_cluster}. \begin{math} \textrm{k-NN}(\mathbf{Z}_{\alpha,p})\end{math} represents
a set of the $\textrm{k}$ nearest, a.k.a, most similar, superpixel representations for $\mathbf{Z}_{\alpha,i}$.
We empirically find that choosing the same neighboring positions as in Equation~\ref{eq:token_cluster}
achieves close performance as the $\textrm{{k}-NN}$ scheme while being more
easy to implementation.

In summary, we estimate
their semantic relations based on the representations before refinement with the following $\beta$ transformer layers and then reconstruct the high-resolution representations from 
the refined low-resolution clustered representations accordingly. 

Finally, we apply the remained $\gamma$ transformer layers to the reconstructed high-resolution features and the task-specific head on the refined high-resolution features to predict the target results such as semantic segmentation maps or monocular depth maps.

\begin{figure*}[t]
\definecolor{mybox1}{RGB}{213,232,212}
\definecolor{mybox2}{RGB}{255,230,204}
\definecolor{mybox3}{RGB}{248,206,204}
\definecolor{mybox4}{RGB}{225,213,231}
\begin{subfigure}[b]{1\textwidth}
\hspace{-5mm}
\includegraphics[width=1.1\textwidth]{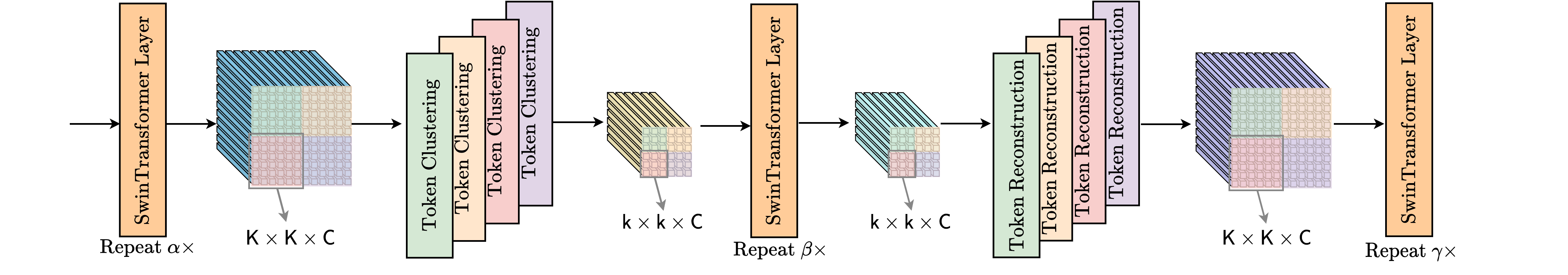}
\end{subfigure}
\caption{Illustrating the details of using our approach for Swin Transformer~\cite{liu2021swinv2,liu2021swinv1}:
we apply four groups of token clustering layer and token reconstruction layer within the four non-overlapped windows marked with \sqbox{mybox1}, \sqbox{mybox2}, \sqbox{mybox3}, and \sqbox{mybox4} respectively, in the example referred to as \emph{window token clustering layer} and \emph{window token reconstruction layer}.
We apply the intermediate swin transformer layers equipped with window size $\mathsf{k}\times \mathsf{k}$) on the clustered window tokens.
Window sizes before and after token clustering layer are $\mathsf{K}\times \mathsf{K}$ and $\mathsf{k}\times \mathsf{k}$, and vice versa, for the token reconstruction layer.
}
\label{fig:swin_cluster_recons}
\vspace{-7mm}
\end{figure*}

\vspace{2mm}
\noindent\textbf{Extension to Swin Transformer.}
We further introduce the \emph{window token clustering layer} and \emph{window token reconstruction layer}, which are suitable for Swin Transformer~\cite{liu2021swinv2,liu2021swinv1}.
Figure~\ref{fig:swin_cluster_recons} illustrates an example usage of the proposed window token clustering layer and window token reconstruction layer. We first cluster the $\mathsf{K}\times \mathsf{K}$ window tokens into $\mathsf{k}\times \mathsf{k}$ window tokens and then reconstruct $\mathsf{K}\times \mathsf{K}$ window tokens according to the refined $\mathsf{k}\times \mathsf{k}$ window tokens.
We apply the swin transformer layer equipped with smaller window size $\mathsf{k}\times \mathsf{k}$ on the clustered representations,
where we need to bi-linear interpolate the pre-trained weights of relative position embedding table from $(2\mathsf{K}-1)$$\times$$(2\mathsf{K}-1)$ to $(2\mathsf{k}-1)$$\times$$(2\mathsf{k}-1)$ when processing the clustered representations.
In summary, we can improve the efficiency of Swin Transformer by injecting the window token clustering layer and the window token reconstruction layer into the backbones seamlessly without fine-tuning the model weights.

\vspace{2mm}
\noindent\textbf{Why our approach can avoid fine-tuning?}
The reasons include the following two aspects:
(i) our token clustering/reconstruction layers are non-parametric, thus avoiding retraining any additional parameters,
(ii) the reconstructed high-resolution representations maintain high semantic similarity with the original high-resolution representations.
We take Segmenter+ViT-L/$16$ (on ADEK, $\alpha$=$10$) as an example and analyze the semantic similarity between the reconstructed high-resolution feature  (with our approach) and the original high-resolution feature (with the original ViT-L/$16$) in Table~\ref{tab:cos_sim}. Accordingly, we can see that the cosine similarities are consistently high across different transformer layers between the reconstructed high-resolution feature (with our approach) and the original high-resolution feature. In other words, our approach well maintains the semantic information carried in the original high-resolution feature maps and thus is capable of avoiding fine-tuning.

\begin{table}[t]
\begin{minipage}[t]{1\linewidth}
\centering
\setlength{\tabcolsep}{12.75pt}
\small
\caption{\small{Ablation of the cosine similarity between the reconstructed high-resolution feature maps and the original high-resolution feature maps, where $\mathbf{Z}_{\alpha+\beta}^{\rm{original}}$ and $\mathbf{Z}_{\alpha+\beta}$ represents the original and reconstructed high-resolution feature maps, respectively.}}
{
\begin{tabular}{ l | c | c | c |  c | c | c | c  }\shline
$\alpha$+$\beta$ & $12$ & $14$ & $16$ & $18$ & $20$ & $22$ & $24$ \\\hline
cos($\mathbf{Z}_{\alpha+\beta}$, $\mathbf{Z}_{\alpha+\beta}^{\rm{original}}$) & $0.94$ & $0.95$ & $0.96$ & $0.96$ & $0.96$ & $0.96$ & $0.96$   \\\hline
\end{tabular}
}
\label{tab:cos_sim}
\end{minipage}
\end{table}

\section{Experiment}
\label{sec:experiment}

We verify the effectiveness of our method across five challenging dense prediction tasks, including object detection, semantic segmentation, instance segmentation, panoptic segmentation, and monocular depth estimation.
We carefully choose the advanced SOTA methods that build the framework based on either the plain ViTs~\cite{dosovitskiy2020image} or the Swin Transformers~\cite{liu2021swinv2,liu2021swinv1}.
We can integrate the proposed token clustering layer and token reconstruction layer seamlessly with the provided official trained checkpoints with no further fine-tuning required.
More experimental details are illustrated as follows.

\subsection{Datasets}

\noindent\textbf{COCO}~\cite{lin2014microsoft}. This dataset consists of $123$K images with $896$K annotated bounding boxes belonging to $80$ thing classes and $53$ stuff classes, where the \texttt{train} set contains $118$K images and the \texttt{val} set contains $5$K images. We report the object detection performance of SwinV2 + HTC++~\cite{liu2021swinv2} and the instance/panoptic segmentation performance of Mask$2$Former~\cite{cheng2021mask2former} on the \texttt{val} set.

\noindent\textbf{ADE$20$K}~\cite{ade20k}. This dataset contains challenging scenes
with fine-grained labels and is one of the most challenging semantic
segmentation datasets. The \texttt{train} set contains $20,210$ images with $150$ semantic classes. The \texttt{val} set contains $2,000$ images.
We report the segmentation results with Segmenter~\cite{Segmenter} on the \texttt{val} set.

\noindent\textbf{PASCAL-Context} \cite{pascal_context}. This dataset consists of $59$ semantic classes plus a background class, where the \texttt{train} set contains $4,996$ images with and the \texttt{val} set contains $5,104$ images. We report the segmentation results with Segmenter~\cite{Segmenter} on the \texttt{val} set.

\noindent\textbf{Cityscapes} \cite{cityscapes}. This dataset is an urban scene understanding dataset with $30$ classes while only $19$ classes are used for parsing evaluation. The \texttt{train} set and \texttt{val} set contains $2,975$ and $500$ images respectively.
We report the segmentation results with Segmenter~\cite{Segmenter} on the \texttt{val} set.

\noindent\textbf{KITTI}~\cite{geiger2012we}. 
This dataset provides stereo, optical flow, visual odometry (SLAM), and 3D object detection of outdoor scenes captured by equipment mounted on a moving vehicle.
We choose DPT~\cite{ranftl2021vision} as our baseline to conduct experiments on the monocular depth prediction tasks, which consists of around $26$K images for \texttt{train} set and $698$ images for \texttt{val} set, where only $653$ images have the ground-truth depth maps and the image resolution is of $1,241\times376$. 

\noindent\textbf{NYUv$2$}~\cite{silberman2012indoor}.
This dataset consists of $1,449$ RGBD images with resolution of $640\times480$, which captures $464$ diverse indoor scenes and contains rich detailed dense annotations such as surface normals, segmentation maps, depth, 3D planes, and so on. We report the depth prediction results of DPT~\cite{ranftl2021vision} evaluated on $655$ \texttt{val} images.

\subsection{Evaluation Metrics}
We report the numbers of AP (average precision), mask AP (mask average precision), PQ (panoptic quality), mIoU (mean intersection-over-union), and RMSE (root mean squared error) across object detection, instance segmentation, panoptic segmentation, semantic segmentation, and depth estimation tasks respectively.
Since the dense prediction tasks care less about the throughput used in image classification tasks \cite{liang2021evit}, we report FPS to measure the latency and the number of GFLOPs to measure the model complexity during evaluation. FPS is tested on a single V100 GPU with Pytorch 1.10 and CUDA 10.2 by default. More details are provided in the supplementary material.

\begin{table}[t]
\begin{minipage}[t]{1\linewidth}
\centering
\setlength{\tabcolsep}{2pt}
\small
\caption{\small{Influence of the hyper-parameters of token clustering/reconstruction layer.}}
{
\begin{tabular}{ l | c | c | c |  c | c | c | c | c | c | c | c | c | c | c }\shline
\multirow{2}{*}{Parameter} & \multicolumn{3}{c|}{$\lambda$} & \multicolumn{4}{c|}{$\kappa$} & \multicolumn{4}{c|}{$\tau$} & \multicolumn{3}{c}{$\mathrm{k}$} \\
\cline{2-15}
&  $3\times3$ & $5\times5$ & $7\times7$ & $5$  & $10$ & $15$ & $20$ & $10$ & $25$ & $50$ & $75$ & $10$ & $20$ & $50$ \\\hline
GFLOPs & $438.6$ & $438.9$ & $439.6$ & $438.9$  & $439.6$ & $440.2$ & $440.8$ & \multicolumn{4}{c|}{$438.9$} & $438.9$ & $438.9$ & $439.0$ \\\hline
mIoU & $51.46$ & $51.56$ & $51.55$ & $51.56$ & $51.59$  & $51.62$ & $51.62$ & $51.18$ & $51.35$ & $51.56$ & $51.07$ & $51.34$ & $51.56$ & $51.56$ \\\shline
\end{tabular}
}
\label{tab:ablation_hyperparam}
\end{minipage}
\begin{minipage}[t]{1\linewidth}
\centering
\setlength{\tabcolsep}{7pt}
\small
\caption{\small{Influence of the cluster size $\mathsf{h}\times\mathsf{w}$.}}
{
\begin{tabular}{ l | c | c | c |  c | c | c | c}\shline
{Cluster size}  & $8\times8$ & $12\times12$ & $16\times16$ & $20\times20$ & $24\times24$ & $28\times28$  & $40\times40$ (baseline)\\\hline
GFLOPs &  $274.0$ & $290.9$ &  $315.1$ & $347.2$  & $388.2$ &  $438.9$ & $659.0$ \\\hline
mIoU & $32.13$ & $44.01$ & $48.21$ &  $50.17$ & $51.32$ & $51.56$ & $51.82$ \\\shline
\end{tabular}
}
\label{tab:ablation_cluster_size}
\end{minipage}
\begin{minipage}[t]{0.3\linewidth}
\centering
\footnotesize
\setlength{\tabcolsep}{1pt}
\caption{\small{Comparison with adaptive average pooling(AAP).}}
\resizebox{\linewidth}{!}
{
\begin{tabular}{ l | c | c }
\shline
\pbox{22mm}{Cluster size}   &  $20\times20$   &  $28\times28$  \\ \hline
AAP           &  $46.45$ &   $46.54$   \\
\rowcolor{cyan!10}Ours          & $50.17$  &   $51.56$   \\\shline
\end{tabular}}
\label{tab:ablation_adative_pool}
\end{minipage}
\begin{minipage}[t]{0.3\linewidth}
\centering
\footnotesize
\setlength{\tabcolsep}{1pt}
\renewcommand{\arraystretch}{1}
\caption{\small{Comparison with bi-linear upsample.}}
\resizebox{\linewidth}{!}
{
\begin{tabular}{ l | c | c }
\shline
\pbox{22mm}{Cluster size}   &  $20\times20$   &  $28\times28$  \\ \hline
Bi-linear           &  $44.68$ &   $44.74$   \\
\rowcolor{cyan!10}Ours                & $50.17$  &   $51.56$  \\\shline
\end{tabular}}
\label{tab:ablation_bilinear}
\end{minipage}
\begin{minipage}[t]{0.4\linewidth}
\centering
\footnotesize
\caption{\small{Combination with lighter vision transformer backbone.}}
\setlength{\tabcolsep}{0.5pt}
\renewcommand{\arraystretch}{1.05}
\resizebox{\linewidth}{!}
{
\begin{tabular}{ l | c | c }
\shline
\pbox{22mm}{Cluster size}   &  mIoU   &  GFLOPs \\ \hline
Segmenter+ViT-B/$16$             &  $48.48$ &  $124.7$   \\
\rowcolor{cyan!10}Segmenter+ViT-B/$16$+Ours       & $48.40$ & $91.9$   \\\shline
\end{tabular}}
\label{tab:ablation_light_backbone}
\end{minipage}
\end{table}

\begin{figure*}[t]
\begin{minipage}[t]{0.5\linewidth}
\hspace{-5mm}
\centering
\begin{subfigure}[b]{0.325\textwidth}
{\includegraphics[width=\textwidth,height=20mm]{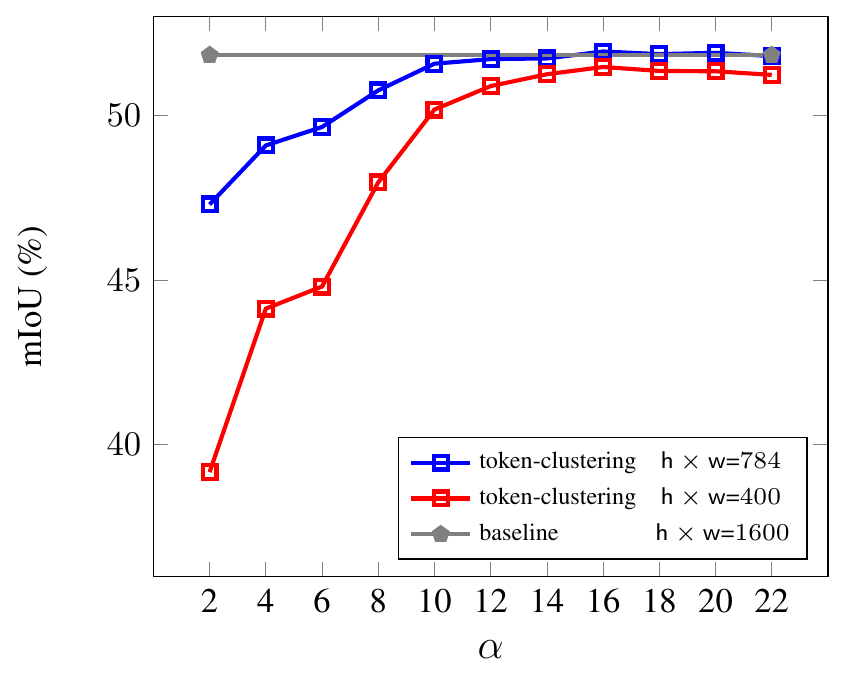}}
\caption{\tiny mIoU}
\end{subfigure}
\begin{subfigure}[b]{0.325\textwidth}
{\includegraphics[width=\textwidth,height=20mm]{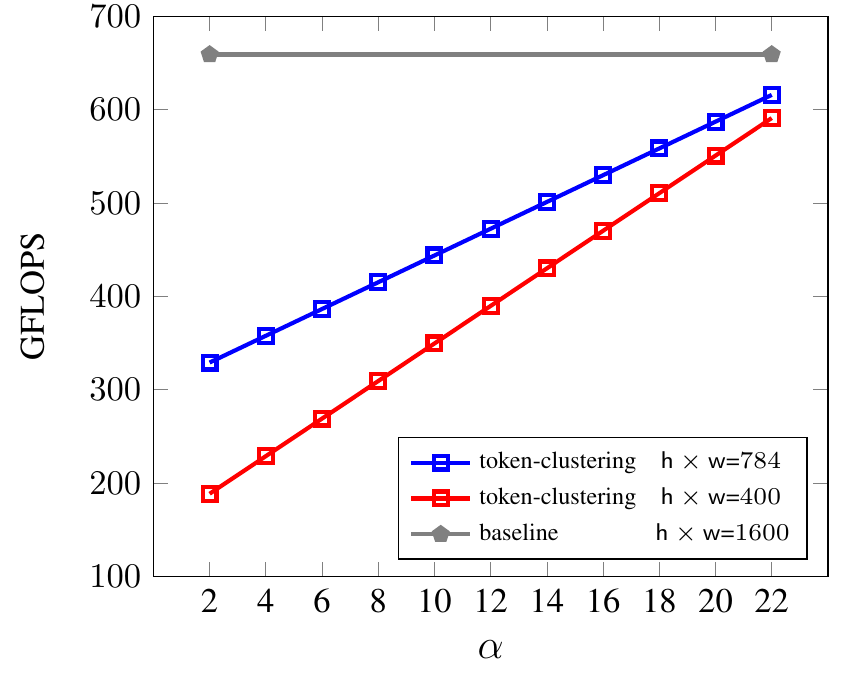}}
\caption{\tiny GFLOPs}
\end{subfigure}
\begin{subfigure}[b]{0.325\textwidth}
{\includegraphics[width=\textwidth,height=20mm]{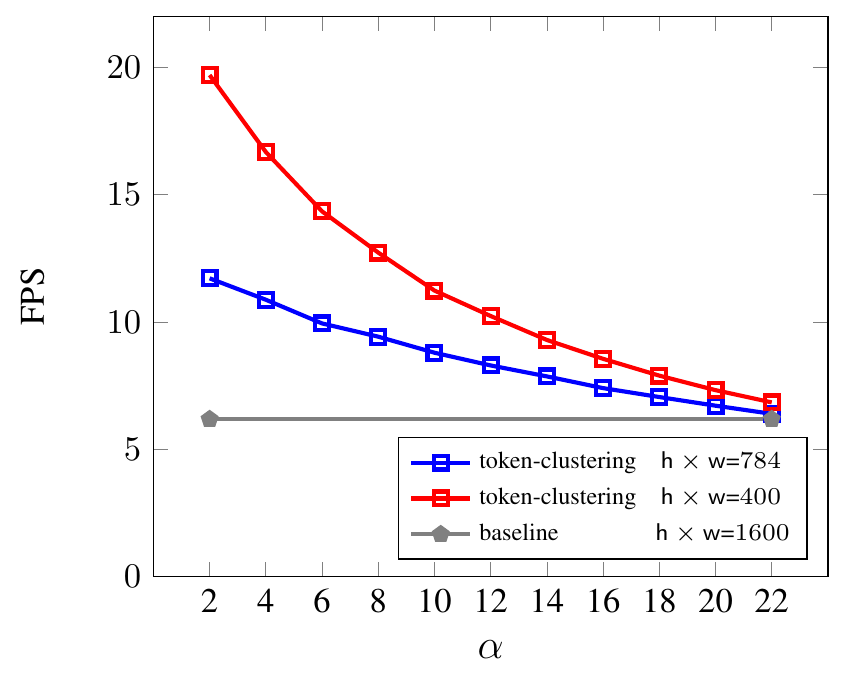}}
\caption{\tiny FPS}
\end{subfigure}
\caption{
\small{Influence of the inserted position $\alpha$ of token clustering layer.}}
\label{fig:ablation_token_cluster_insert_position}
\end{minipage}
\begin{minipage}[t]{0.5\linewidth}
\hspace{-2mm}
\centering
\begin{subfigure}[b]{0.325\textwidth}
{\includegraphics[width=\textwidth,height=20mm]{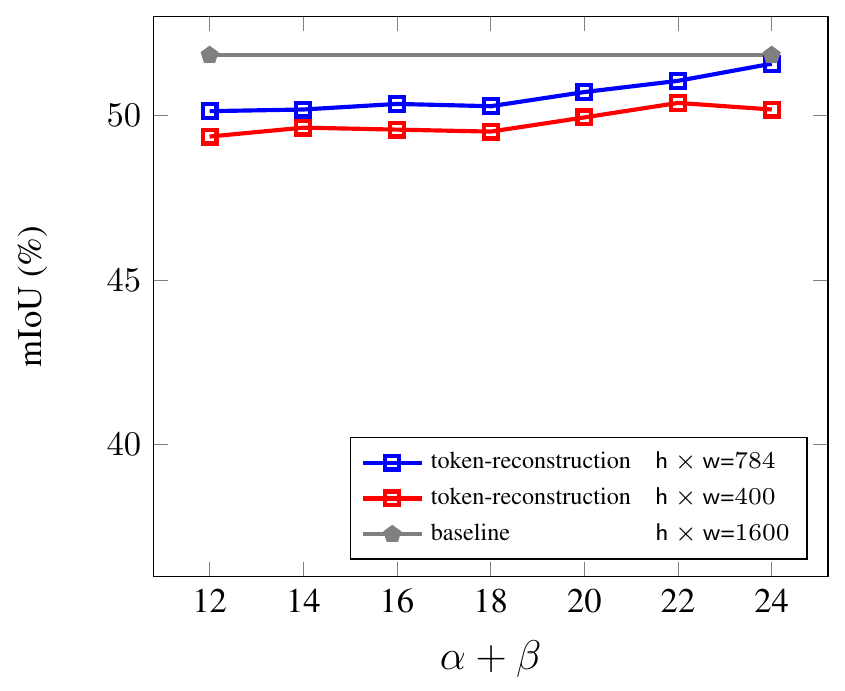}}
\caption{\tiny mIoU}
\end{subfigure}
\begin{subfigure}[b]{0.325\textwidth}
{\includegraphics[width=\textwidth,height=20mm]{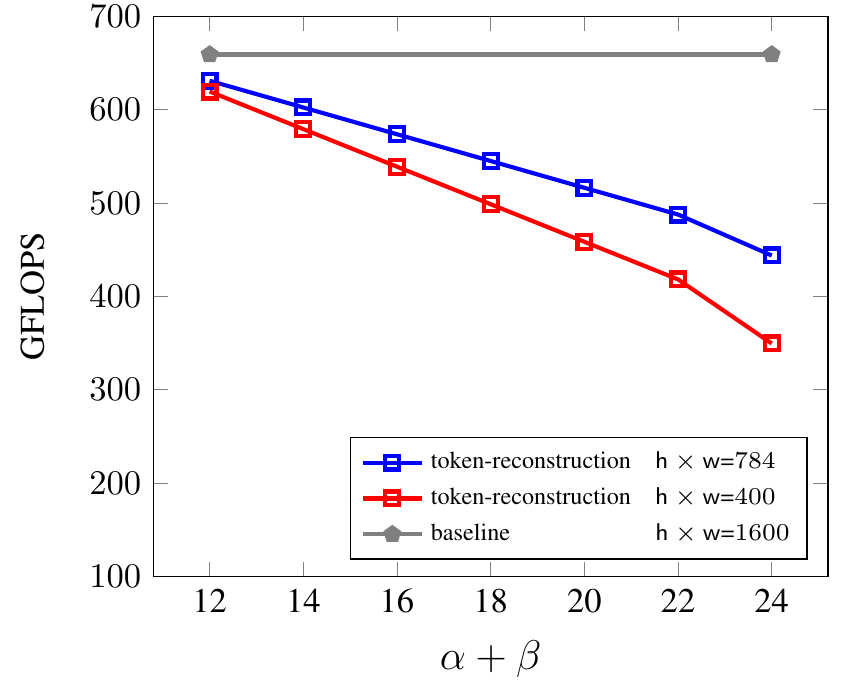}}
\caption{\tiny GFLOPs}
\end{subfigure}
\begin{subfigure}[b]{0.325\textwidth}
{\includegraphics[width=\textwidth,height=20mm]{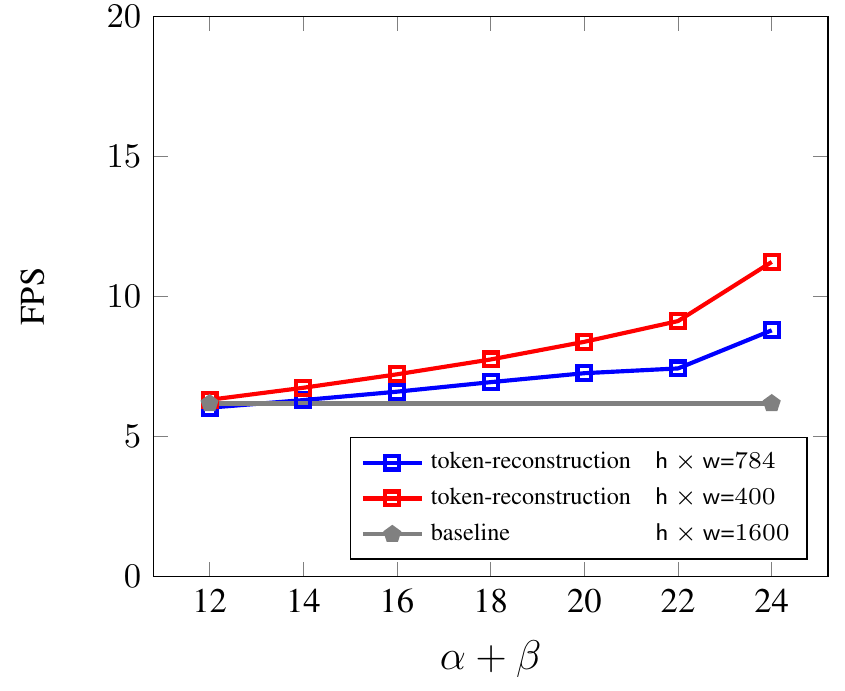}}
\caption{\tiny FPS}
\end{subfigure}
\caption{
\small{Influence of the inserted position $\alpha+\beta$ of token reconstruction layer.}}
\label{fig:ablation_token_reconstruct_insert_position}
\end{minipage}
\end{figure*}

\subsection{Ablation Study Experiments}
We conduct the following ablation experiments on ADE$20$K semantic segmentation benchmark with the official checkpoints of Segmenter+ViT-L/$16$~\cite{Segmenter}\footnote{\paperurl{{https://github.com/rstrudel/segmenter##ade20k}}, {MIT License}} by default if not specified.

\noindent\textbf{Hyper-parameters of token clustering/reconstruction layer}.
We first study the influence of the hyper-parameters associated with the token clustering layer, i.e., the number of neighboring pixels $\lambda$ used in Equation~\ref{eq:token_cluster}, the number of EM iterations $\kappa$, and the choice of the temperature $\tau$ in Table~\ref{tab:ablation_hyperparam}. According to the results, we can see that our method is relatively less sensitive to the choice of both $\lambda$ and $\kappa$ compared to $\tau$. In summary, we choose $\lambda$ as $5\times5$, $\kappa$ as $5$, and $\tau$ as $50$ considering both performance and efficiency.
Next, we also study the influence of the hyper-parameters within the token clustering layer, i.e., the number of nearest neighbors $\textrm{k}$ within $\textrm{k-NN}$.
We do not observe obvious differences and thus set $\textrm{k}$ as $20$. More details are provided in the supplementary material.

\noindent\textbf{Influence of cluster size choices}.
We study the influence of different cluster size $\mathsf{h}\times\mathsf{w}$ choices based on input feature map of size $\frac{\mathsf{H}}{\mathsf{P}}\times\frac{\mathsf{W}}{\mathsf{P}}=40\times40$ ($\mathsf{N}=1,600$)\footnote{{We choose $\mathsf{H}\times\mathsf{W}=640\times640$ and $\mathsf{P}=16$, thus, $\frac{\mathsf{H}}{\mathsf{P}}\times\frac{\mathsf{W}}{\mathsf{P}}=40\times40$ or $\mathsf{N}=1,600$, on ADE$20$K.}} in Table~\ref{tab:ablation_cluster_size}.
According to the results, we can see that choosing too small cluster sizes significantly harms the dense prediction performance, and setting $\mathsf{h}\times\mathsf{w}$ as $28\times28$ achieves
the better trade-off between performance drop and
model complexity.
Therefore, we choose $28\times28$ on Segmenter+ViT-L/$16$ by default.
We also empirically find that selecting the cluster size $\mathsf{h}\times\mathsf{w}$
around $\mathsf{N}/4\sim\mathsf{N}/2$ performs better on most of the other experiments.
We conduct the following ablation experiments under two typical settings, including $20\times20$ ($\sim\mathsf{N}/4$) and $28\times28$ ($\sim\mathsf{N}/2$).

\noindent\textbf{Comparison with adaptive average pooling and bi-linear upsample}.
We report the comparison results between our proposed token clustering/reconstruction scheme and adaptive average pooling/bi-linear upsample scheme in Table~\ref{tab:ablation_adative_pool} and Table~\ref{tab:ablation_bilinear} under two cluster size settings respectively.
We choose to compare with adaptive average pooling and bi-linear upsample instead of strided convolution or deconvolution as the previous ones are non-parametric and the later ones require re-training or fine-tuning, which are not the focus of this work.
Specifically, we keep the inserted position choices the same and only replace the token cluster or token reconstruction layer with adaptive average pooling or bi-linear upsampling under the same cluster size choices. According to the results, we can see that our proposed token clustering and token reconstruction consistently outperform adaptive average pooling and bi-linear upsampling under different cluster size choices.

\noindent\textbf{Influence of inserted position of token clustering/reconstruction layer}.
We investigate the influence of the inserted position of both token clustering layer and token reconstruction layer and summarize the detailed results in Figure~\ref{fig:ablation_token_cluster_insert_position} and Figure~\ref{fig:ablation_token_reconstruct_insert_position} under two different cluster size choices.
According to the results shown in Figure~\ref{fig:ablation_token_cluster_insert_position},
our method achieves better performance when choosing $\alpha$ larger than $10$,
therefore, we choose $\alpha=10$ as it achieves a better trade-off between model complexity and segmentation accuracy.
Then we study the influence of the inserted positions of the token reconstruction layer
by fixing $\alpha=10$.
According to Figure~\ref{fig:ablation_token_reconstruct_insert_position},
we can see that our method achieves the best performance when setting $\alpha+\beta=24$,
in other words, we insert the token reconstruction layer after the last transformer layer of ViT-L/$16$.
We choose $\alpha=10$, $\alpha+\beta=24$, and $\gamma=0$ for all ablation experiments on ADE$20$K by default if not specified.

\noindent\textbf{Combination with lighter vision transformer architecture}.
We report the results of applying our method to lighter vision transformer backbones such as ViT-B/$16$, in Table~\ref{tab:ablation_light_backbone}.
Our approach consistently improves the efficiency of Segmenter+ViT-B/$16$
at the cost of a slight performance drop without fine-tuning.
Specifically speaking, our approach saves more than $26\%\downarrow$ GFLOPs of a trained ``Segmenter+ViT-B/$16$'' with only a slight performance drop from $48.48\%$ to $48.40\%$, which verifies our method also generalizes to lighter vision transformer architectures.

\noindent\textbf{Comparison with uniform downsampling}.
We compare our method with the simple uniform downsampling scheme, which directly downsamples the input image into a lower resolution.
Figure~\ref{fig:comp_uniform_downsample} summarizes the detailed comparison results.
For example, on ADE$20$K, we downsample the input resolution from $640\times640$ to smaller resolutions (\eg, $592\times592$, $576\times576$, $560\times560$, and $544\times544$) and report their performance and GFLOPs in Figure~\ref{fig:comp_uniform_downsample}.
We also plot the results with our method and we can see that our method consistently outperforms uniform sampling on both ADE$20$K and PASCAL-Context
under multiple different GFLOPs budgets.

\subsection{Object Detection}

\begin{figure*}[t]
\begin{minipage}[t]{0.395\linewidth}
\centering
\hspace{-3mm}
\begin{subfigure}[b]{0.485\textwidth}
{\includegraphics[width=\textwidth,height=25mm]{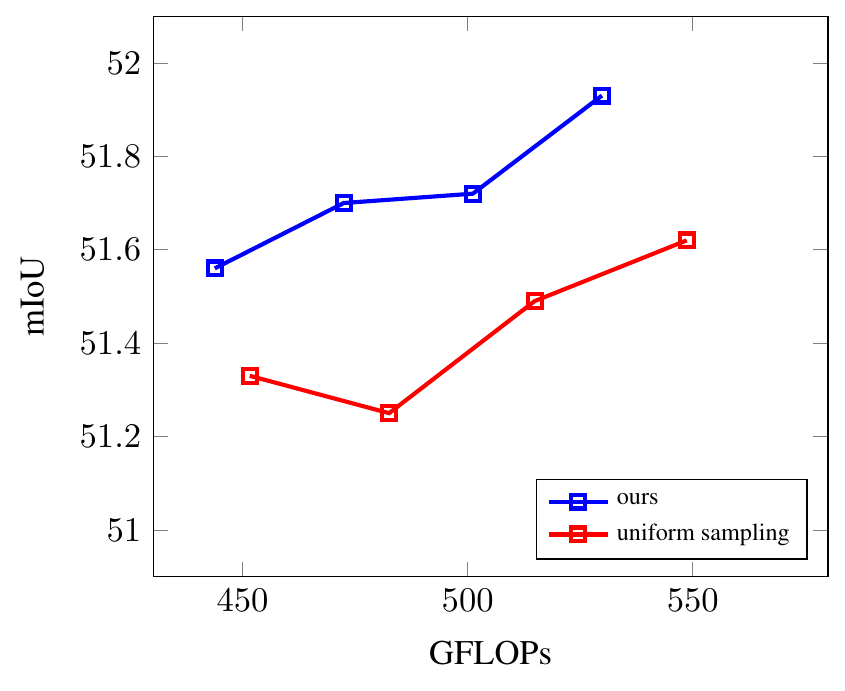}}
\caption{\tiny ADE$20$K}
\end{subfigure}
\begin{subfigure}[b]{0.485\textwidth}
{\includegraphics[width=\textwidth,height=25mm]{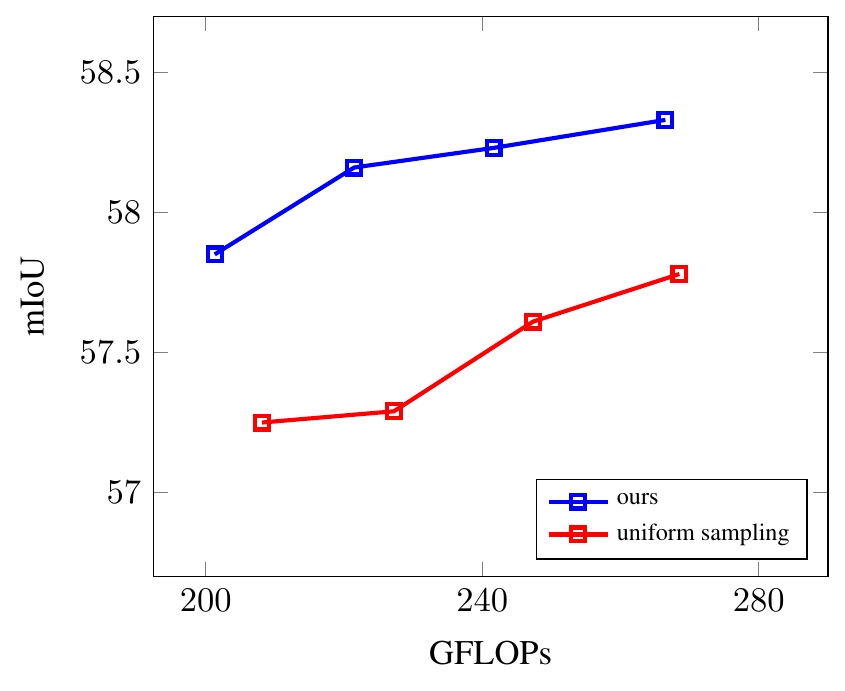}}
\caption{\tiny PASCAL-Context}
\end{subfigure}
\caption{
\small{Comparison to uniform downsampling on ADE$20$K and PASCAL-Context with Segmenter+ViT-L/$16$.}}
\label{fig:comp_uniform_downsample}
\end{minipage}
\hspace{1mm}
\begin{minipage}[t]{0.6\linewidth}
\centering
\begin{subfigure}[b]{0.47\textwidth}
\hspace{-3mm}
{\includegraphics[width=\textwidth,height=25mm]{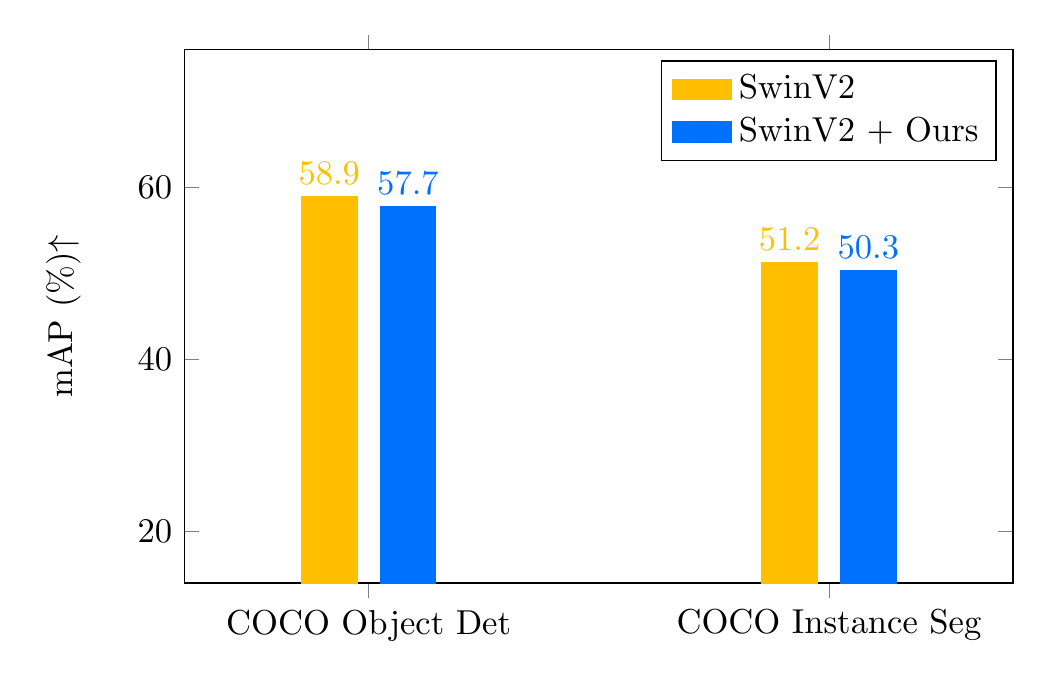}}
\caption{\tiny{mAP}}
\end{subfigure}
\begin{subfigure}[b]{0.255\textwidth}
{\includegraphics[width=\textwidth,height=25mm]{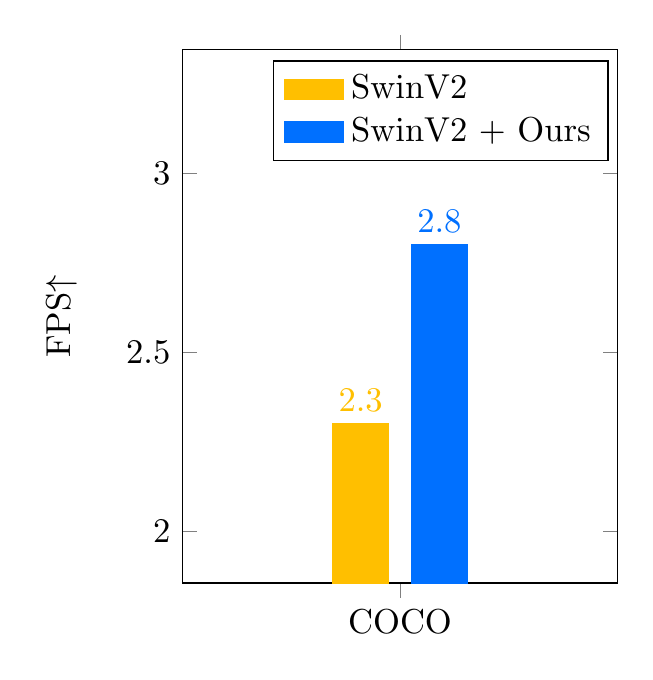}}
\caption{\tiny{FPS}}
\end{subfigure}
\begin{subfigure}[b]{0.255\textwidth}
{\includegraphics[width=\textwidth,height=25mm]{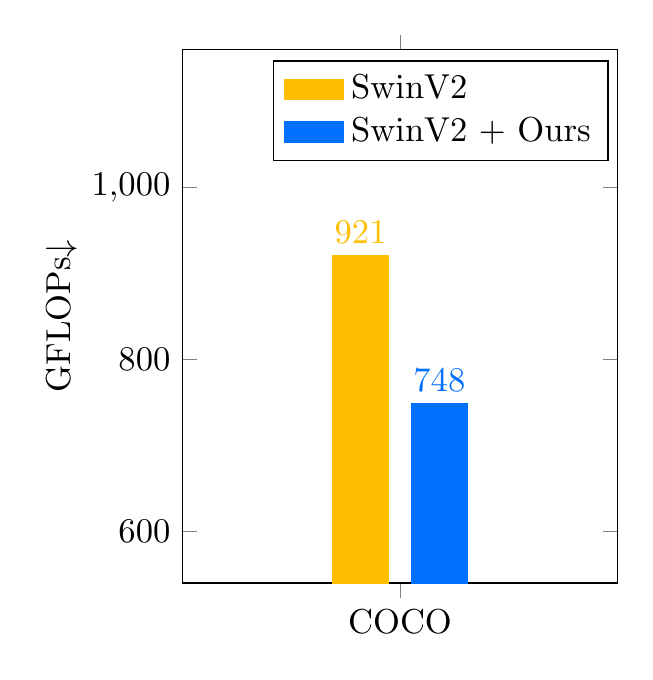}}
\caption{\tiny{GFLOPs}}
\end{subfigure}
\caption{\small{Illustrating the improvements of our approach on object detection task with SwinV$2$-L + HTC++. $\uparrow$ and $\downarrow$ represent higher is better and lower is better respectively.}}
\label{fig:swinv2_result}
\end{minipage}
\begin{minipage}[t]{1\linewidth}
\centering
\begin{subfigure}[b]{0.325\textwidth}
\hspace{-3mm}
{\includegraphics[width=\textwidth,height=25mm]{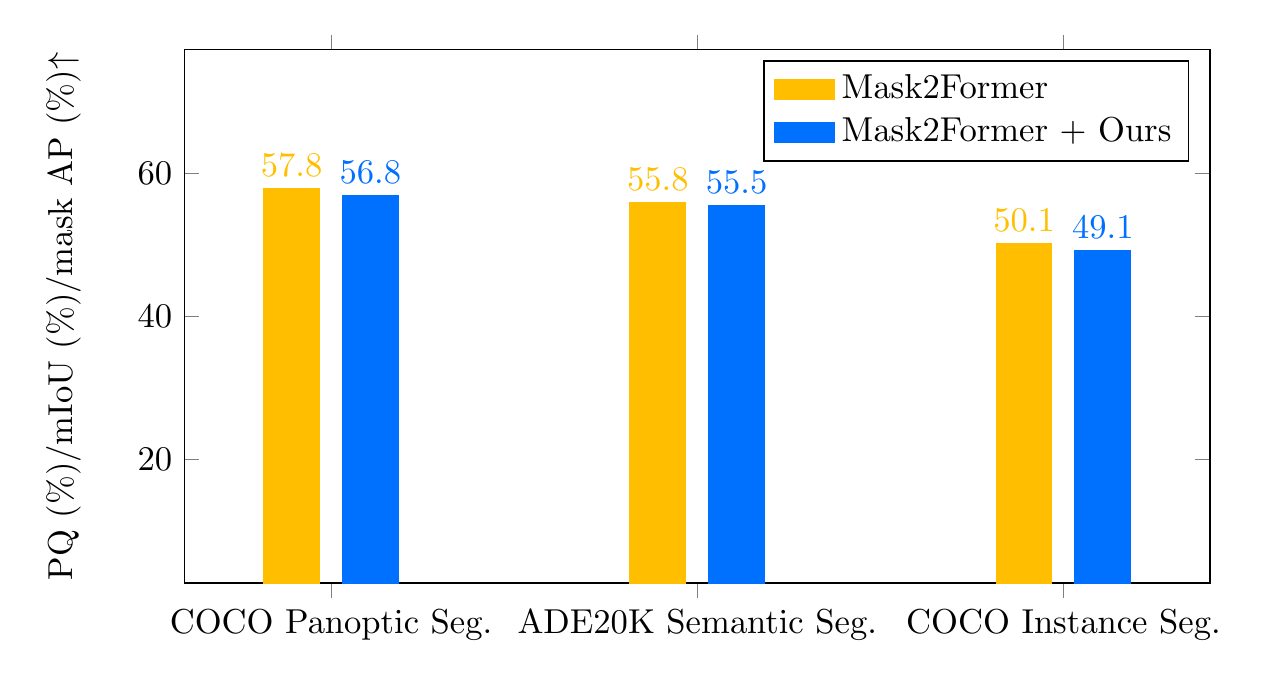}}
\caption{\tiny{mAP}}
\end{subfigure}
\begin{subfigure}[b]{0.325\textwidth}
{\includegraphics[width=\textwidth,height=25mm]{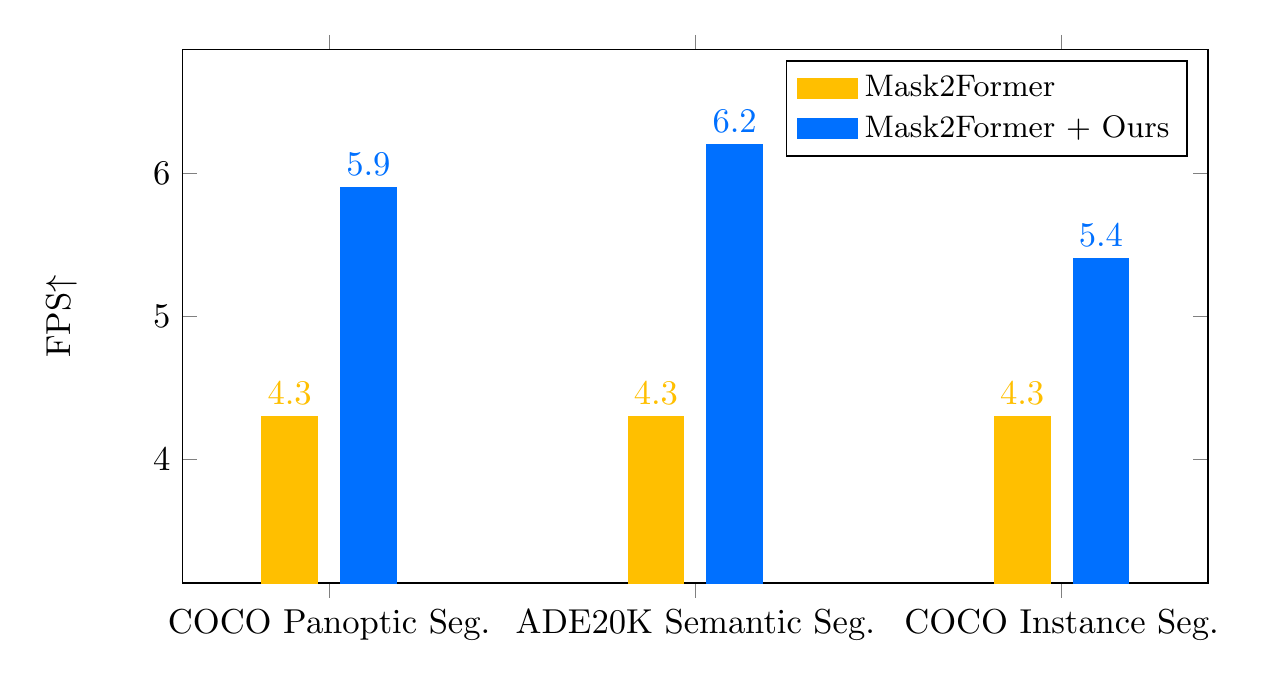}}
\caption{\tiny{FPS}}
\end{subfigure}
\begin{subfigure}[b]{0.325\textwidth}
{\includegraphics[width=\textwidth,height=25mm]{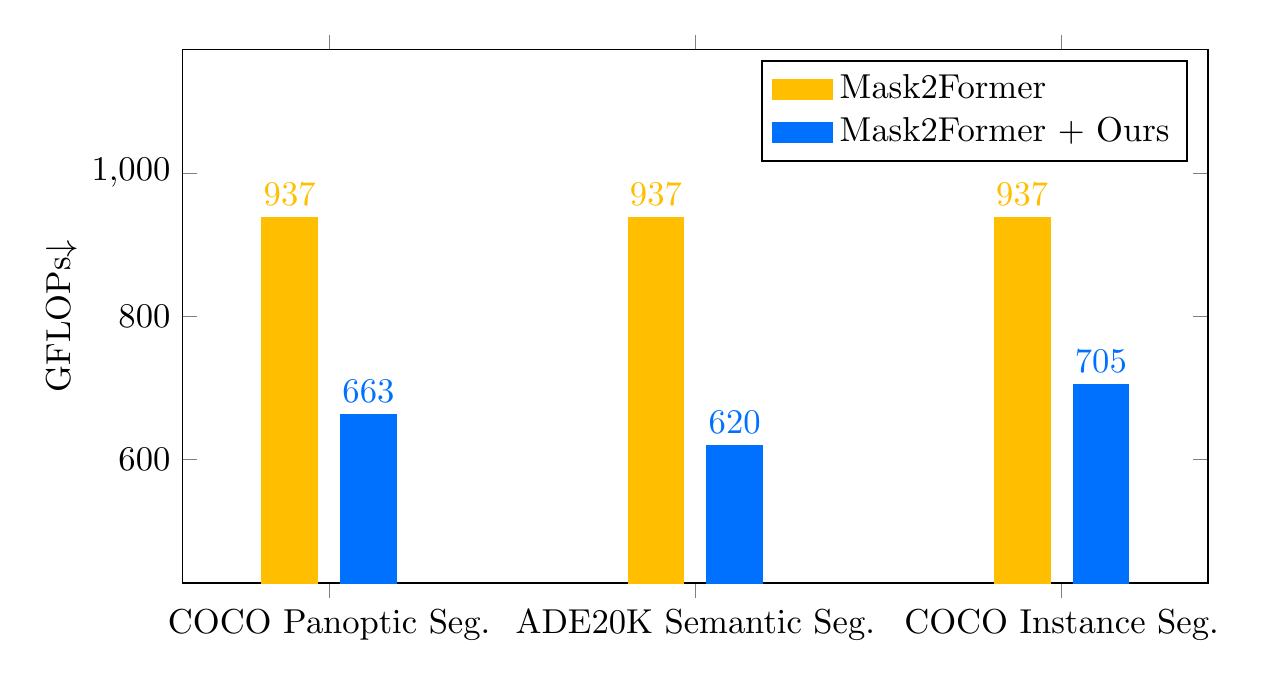}}
\caption{\tiny{GFLOPs}}
\end{subfigure}
\caption{\small{Illustrating the improvements of our approach on panoptic segmentation, semantic segmentation, and instance segmentation tasks with Mask$2$Former+Swin-L.}}
\label{fig:mask2former_result}
\end{minipage}
\begin{minipage}[t]{1\linewidth}
\centering
\begin{subfigure}[b]{0.325\textwidth}
\hspace{-3mm}
{\includegraphics[width=\textwidth,height=25mm]{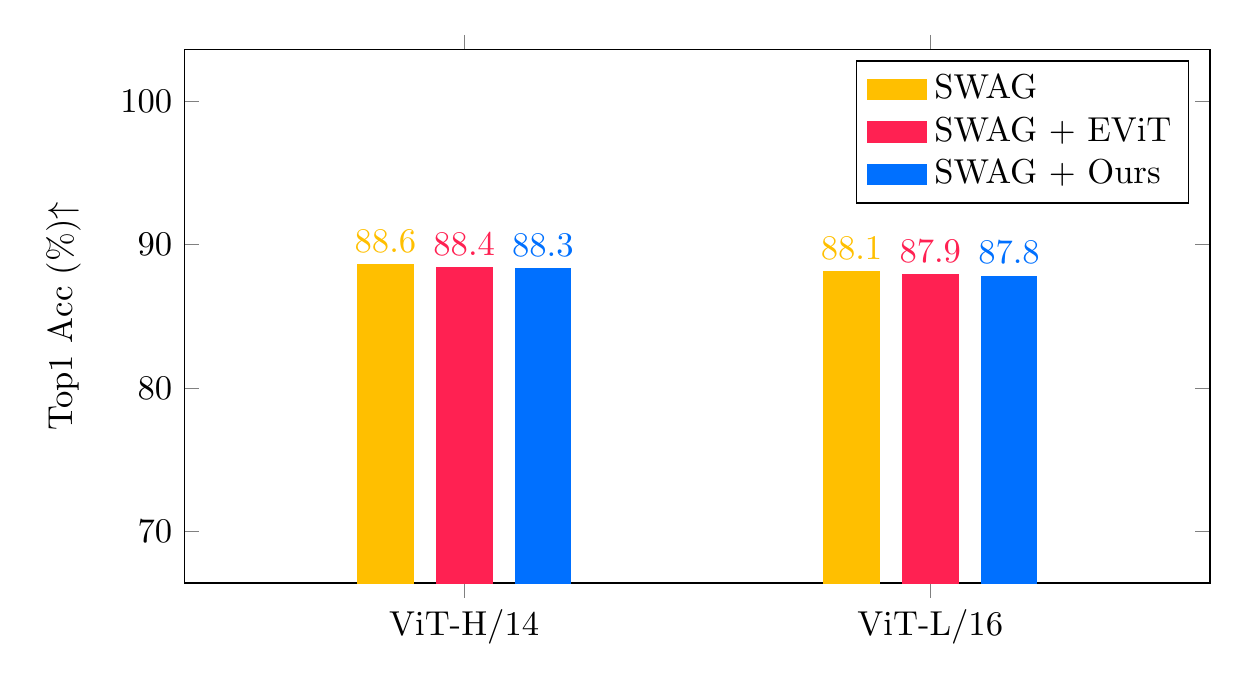}}
\caption{\tiny{Top-$1$ Accuracy}}
\end{subfigure}
\begin{subfigure}[b]{0.325\textwidth}
{\includegraphics[width=\textwidth,height=25mm]{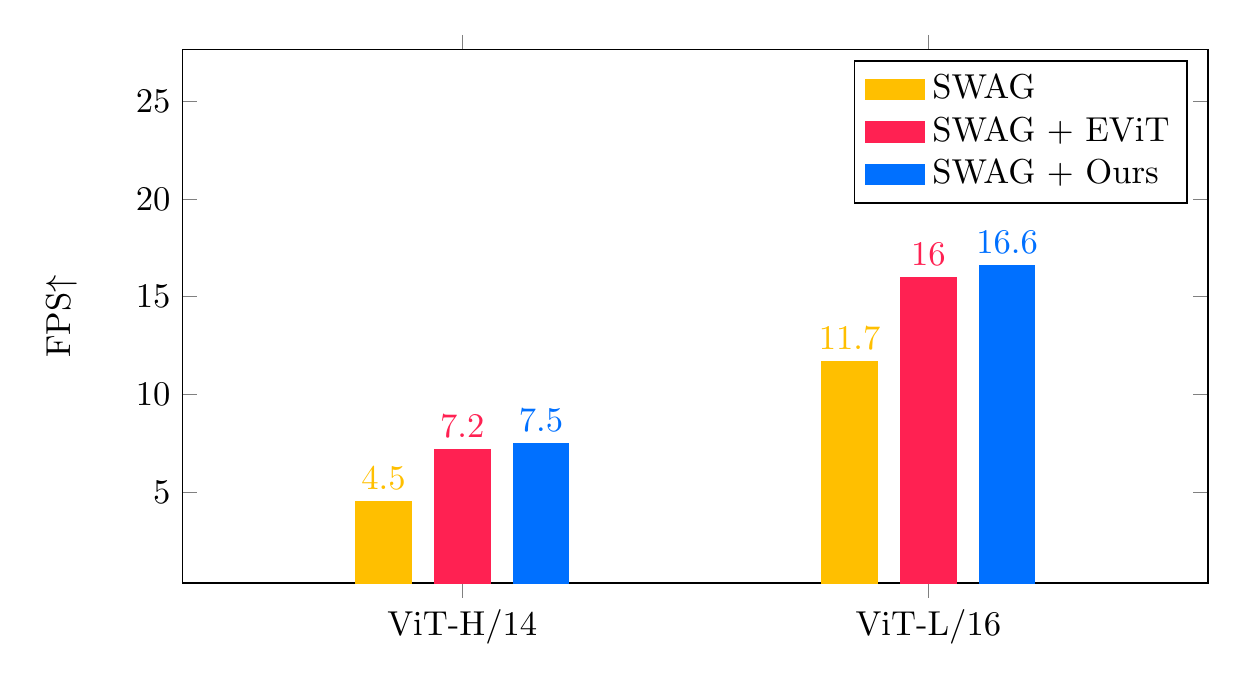}}
\caption{\tiny{FPS}}
\end{subfigure}
\begin{subfigure}[b]{0.325\textwidth}
{\includegraphics[width=\textwidth,height=25mm]{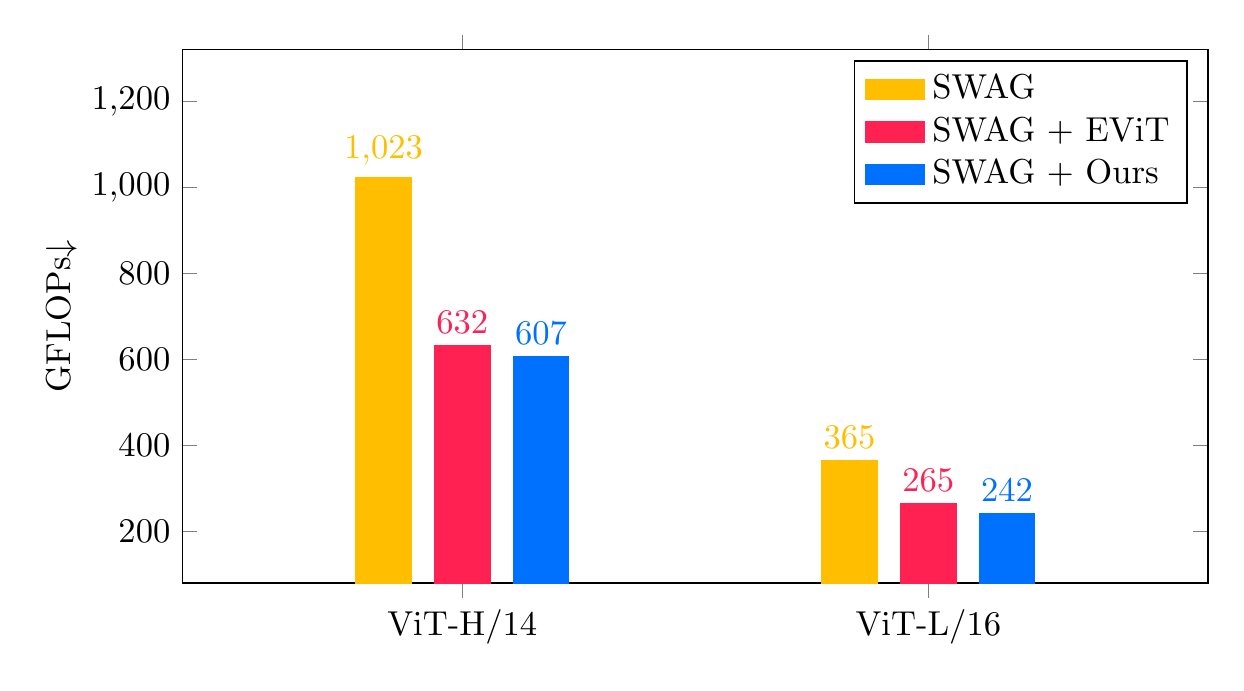}}
\caption{\tiny{GFLOPs}}
\end{subfigure}
\caption{\small{Illustrating the improvements of our approach and the comparisons with EViT~\cite{liang2021evit} on ImageNet classification tasks with SWAG+ViT-H/$14$ and SWAG+ViT-L/$16$.}}
\label{fig:swag_result}
\end{minipage}
\vspace{-5mm}
\end{figure*}

We use the recent SOTA object detection framework SwinV$2$-L + HTC++~\cite{liu2021swinv2} as our baseline.
We summarize the results of combining our method with SwinV$2$-L + HTC++ on COCO object detection and instance segmentation tasks in Figure~\ref{fig:swinv2_result}.

\noindent\textbf{Implementation details.}
The original SwinV$2$-L consists of \{$2$,$2$,$18$,$2$\} shifted window transformer blocks across the four stages.
We only apply our method to the $3$-rd stage with $18$ blocks considering it dominates the computation overhead, which we also follow in the ``Mask$2$Former + Swin-L'' experiments.
We insert the window token clustering/reconstruction layer after the $8$-th/$18$-th block within the $3$-rd stage, which are based on the official checkpoints~\footnote{\paperurl{https://github.com/microsoft/Swin-Transformer}, {MIT License}} of SwinV$2$-L + HTC++.
In other words, we set $\alpha=12$ and $\alpha+\beta=22$ for SwinV$2$-L.
The default window size is $\mathsf{K}\times\mathsf{K}$=$32\times32$ and we set the clustered window size as $\mathsf{k}\times\mathsf{k}$=$23\times23$.
We choose the values of other hyperparameters following the ablation experiments.
According to the results summarized in Figure~\ref{fig:swinv2_result},
compared to SwinV$2$-L + HTC++,
our method improves the FPS by {$21\%\uparrow$} and saves the GFLOPs by nearly {$20\%\downarrow$} while maintaining around $98\%$ of object detection \& instance segmentation performance.

\subsection{Semantic/Instance/Panoptic Segmentation}

We first apply our method to a plain ViT-based segmentation framework Segmenter~\cite{Segmenter} and illustrate the semantic segmentation results across three benchmarks including ADE$20$K, PASCAL-Context, and Cityscapes on the first row of Figure~\ref{fig:intro_results}. Then, we apply our method to a very recent framework Mask$2$Former~\cite{cheng2021mask2former} that is based on Swin Transformer and summarize the semantic segmentation, instance segmentation, and panoptic segmentation results on COCO in Figure~\ref{fig:mask2former_result}.

\noindent\textbf{Implementation details.}
The original ViT-L first splits an image into a sequence of image patches of size $16\times16$ and applies a patch embedding layer to increase the channel dimensions to $1024$, then applies $24$ consecutive transformer encoder layers for representation learning. To apply our method to the ViT-L backbone of Segmenter, we use the official checkpoints~\footnote{\paperurl{https://github.com/rstrudel/segmenter##model-zoo}, {MIT License}} of ``Segmenter + ViT-L/$16$'' and insert the token clustering layers and token reconstruction layer into the ViT-L/$16$ backbone without fine-tuning.

For the Mask$2$Former built on Swin-L with window size as $12\times12$, we use the official checkpoints~\footnote{\paperurl{https://github.com/facebookresearch/Mask2Former/blob/main/MODEL_ZOO.md}, {CC-BY-NC $4.0$}} of ``Mask$2$Former + Swin-L'' and insert the window token clustering layer and the window token reconstruction layer into the empirically chosen positions,
which first cluster $12\times12$ tokens into $8\times8$ tokens and then reconstruct $12\times12$ tokens within each window. Figure~\ref{fig:mask2former_result} summarizes the detailed comparison results. Accordingly, we can see that our method significantly improves the FPS by more than $35\%\uparrow$ with a slight performance drop on COCO panoptic segmentation task.

\subsection{Monocular Depth Estimation}
To verify the generalization of our approach, we apply our method to depth estimation tasks that measure the distance of each pixel relative to the camera. We choose the DPT (Dense Prediction Transformer)~\cite{ranftl2021vision} that builds on the hybrid vision transformer, i.e., R$50$+ViT-B/$16$, following~\cite{dosovitskiy2020image}.

\noindent\textbf{Implementation details.}
The original R$50$+ViT-B/$16$~\footnote{\paperurl{https://github.com/isl-org/DPT}, {MIT License}} consists of a ResNet$50$ followed by a ViT-B/$16$,
where the ViT-B/$16$ consists of $12$ transformer encoder layers that process $16\times$ downsampled representations.
We insert the token clustering layer \& token reconstruction layer into ViT-B/$16$ and summarize the results on both KITTI and NYUv$2$ on the second row of Figure~\ref{fig:intro_results}. We also report their detailed depth estimation results in Table~\ref{table:dpt_result},
where we can see that our method accelerates DPT by nearly $30\%$/$37\%\uparrow$ on KITTI/NYUv$2$, respectively.

\subsection{ImageNet-$1$K Classification}
Finally, we apply our method to the ImageNet-$1$K classification task and
compare our method with a very recent SOTA method EViT~\cite{liang2021evit}.
The key idea of EViT is to identify and only keep the top-$k$ tokens according to their
attention scores relative to the [\texttt{class}] token.
We empirically find that applying EViT for dense prediction tasks directly
suffers from significant performance drops. More details are illustrated in the supplementary material.

\noindent\textbf{Implementation details.}
We choose the recent SWAG~\cite{singh2022revisiting} as our baseline,
which exploits $3.6$ billion weakly labeled images associated with around $27$K categories (or hashtags)
to pre-train the large-scale vision transformer models, i.e., ViT-L/$16$ and ViT-H/$14$.
According to their official implementations~\footnote{\paperurl{https://github.com/facebookresearch/SWAG}, {CC-BY-NC $4.0$}},
SWAG + ViT-H/$14$ and SWAG + ViT-L/$16$ achieve $88.55\%$ and $88.07\%$ top-$1$ accuracy on ImaegNet-1K respectively.
We apply our approach and EViT to both baselines and summarize the comparison results in Figure~\ref{fig:swag_result}.
According to the results,
our method achieves comparable results as EViT while being more efficient, which further
verifies that our method also generalizes to the image classification tasks without fine-tuning.

\begin{table}[t]
\begin{minipage}[t]{\linewidth}
\centering
\footnotesize
\caption{\small{Depth estimation results based on DPT~\cite{ranftl2021vision} with ResNet-$50$+ViT-B/$16$.}}
\setlength\tabcolsep{1.25pt}
\resizebox{\linewidth}{!}{
\begin{tabular}{l|l|cc|ccccccccc}
\shline
Dataset & Method  & GFLOPs & FPS & {$\delta$>$1.25$} & {$\delta$>$1.25^{2}$} & {$\delta$>$1.25^{3}$} & AbsRel  & SqRel & RMSE & RMSElog & SILog  & {log$10$}\\
\hline
\multirow{2}{*}{KITTI} & {DPT} & $810$ & $11.38$ &  $0.959$ & $0.995$  & $0.999$ & $0.062$ & $0.222$ &  $2.573$ & $0.092$   &  $8.282$ & $0.027$\\
& \cellcolor{cyan!10}{DPT+Ours} & \cellcolor{cyan!10}$627$ &\cellcolor{cyan!10}\cellcolor{cyan!10}$14.75$ &  \cellcolor{cyan!10}$0.958$ & \cellcolor{cyan!10}$0.995$ &\cellcolor{cyan!10}$0.999$   & \cellcolor{cyan!10}$0.062$   & \cellcolor{cyan!10}$0.226$   &  \cellcolor{cyan!10}$2.597$ &\cellcolor{cyan!10}$0.093$   & \cellcolor{cyan!10}$8.341$ & \cellcolor{cyan!10}$0.027$\\
\hline
\multirow{2}{*}{NYUv$2$} & {DPT} & $560$ & $17.58$  &  $0.904$    & $0.988$  & $0.998$   & $0.110$ & $0.054$   &  $0.357$ & $0.129$   &  $9.522$ & $0.045$\\
& \cellcolor{cyan!10}{DPT+Ours} &\cellcolor{cyan!10}$404$ &\cellcolor{cyan!10}$24.03$ &\cellcolor{cyan!10}$0.900$   &\cellcolor{cyan!10}$0.987$   &\cellcolor{cyan!10}$0.998$	&\cellcolor{cyan!10}$0.113$ &\cellcolor{cyan!10}$0.056$ &\cellcolor{cyan!10}$0.363$ &\cellcolor{cyan!10}$0.132$ &\cellcolor{cyan!10}$9.532$ &\cellcolor{cyan!10}$0.046$\\
\shline
\end{tabular}
}
\label{table:dpt_result}
\end{minipage}
\end{table}

\section{Conclusion}
In this paper, we present a simple and effective mechanism to improve the efficiency of large-scale vision transformer models for dense prediction tasks. In light of the relatively high costs associated with re-training or fine-tuning large vision transformer models on various dense prediction tasks, our study provides a very lightweight method for expediting the inference process while requiring no additional fine-tuning.
We hope our work could inspire further research efforts into exploring how to accelerate large-scale vision transformers for dense prediction tasks without fine-tuning.

\section*{Acknowledgement}
This work is partially supported by the National Nature Science Foundation of China under Grant $62071013$ and $61671027$, and National Key R\&D Program of China under Grant $2018$AAA$0100300$.

\bibliography{egbib}
\bibliographystyle{plain}

\clearpage
\newpage

\section*{Checklist}

\begin{enumerate}

\item For all authors...
\begin{enumerate}
  \item Do the main claims made in the abstract and introduction accurately reflect the paper's contributions and scope?
    \answerYes[]
  \item Have you read the ethics review guidelines and ensured that your paper conforms to them?
    \answerYes[]
  \item Did you discuss any potential negative societal impacts of your work?
    \answerNA{}
  \item Did you describe the limitations of your work?
    \answerNA[]
\end{enumerate}

\item If you are including theoretical results...
\begin{enumerate}
  \item Did you state the full set of assumptions of all theoretical results?
    \answerNA{}
	\item Did you include complete proofs of all theoretical results?
    \answerNA{}
\end{enumerate}

\item If you ran experiments...
\begin{enumerate}
  \item Did you include the code, data, and instructions needed to reproduce the main experimental results (either in the supplemental material or as a URL)?
    \answerTODO{We will release the code soon.}
  \item Did you specify all the training details (e.g., data splits, hyperparameters, how they were chosen)?
    \answerYes{}
	\item Did you report error bars (e.g., with respect to the random seed after running experiments multiple times)?
    \answerTODO{}
	\item Did you include the total amount of compute and the type of resources used (e.g., type of GPUs, internal cluster, or cloud provider)?
    \answerYes{}
\end{enumerate}

\item If you are using existing assets (e.g., code, data, models) or curating/releasing new assets...
\begin{enumerate}
  \item If your work uses existing assets, did you cite the creators?
    \answerYes{We use the official checkpoints provided by:
    \begin{itemize}
    \item \paperurl{{https://github.com/rstrudel/segmenter\#model-zoo}}, MIT License
    \item \paperurl{https://github.com/facebookresearch/Mask2Former/blob/main/MODEL_ZOO.md}, CC-BY-NC $4.0$
    \item \paperurl{https://github.com/microsoft/Swin-Transformer}, MIT License
    \item \paperurl{https://github.com/facebookresearch/SWAG}, CC-BY-NC $4.0$
    \item \paperurl{https://github.com/isl-org/DPT}, MIT License.
    \end{itemize}
    }
  \item Did you mention the license of the assets?
    \answerYes{We mark the license of these assets as above.}
  \item Did you include any new assets either in the supplemental material or as a URL?
    \answerYes{We add the URL of these assets in the footnote.}
  \item Did you discuss whether and how consent was obtained from people whose data you're using/curating?
    \answerNA{}
  \item Did you discuss whether the data you are using/curating contains personally identifiable information or offensive content?
    \answerNA{}
\end{enumerate}

\item If you used crowdsourcing or conducted research with human subjects...
\begin{enumerate}
  \item Did you include the full text of instructions given to participants and screenshots, if applicable?
    \answerNA{}
  \item Did you describe any potential participant risks, with links to Institutional Review Board (IRB) approvals, if applicable?
    \answerNA{}
  \item Did you include the estimated hourly wage paid to participants and the total amount spent on participant compensation?
    \answerNA{}
\end{enumerate}

\end{enumerate}

\newpage
\section*{Appendix}
\label{sec:appendix}

\begin{figure*}[h]
\centering
{\includegraphics[width=\textwidth]{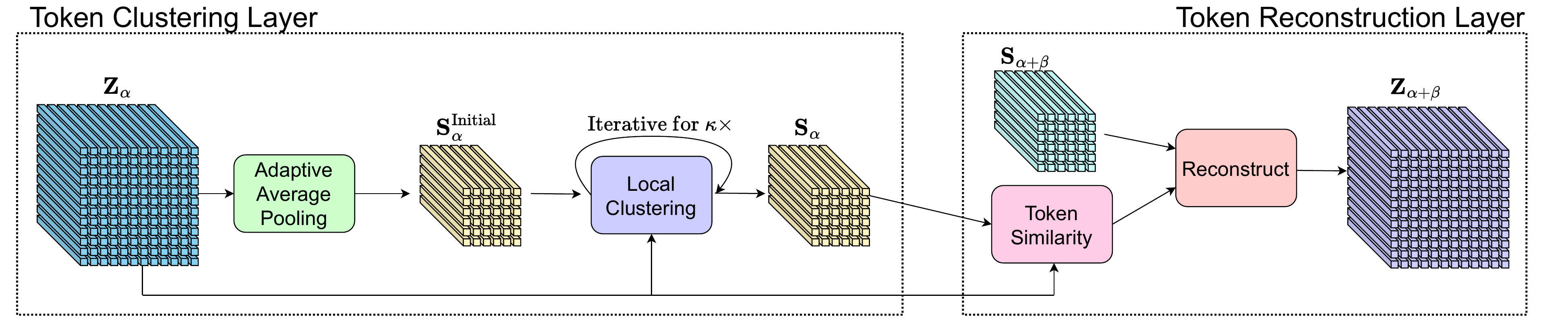}}
\caption{\small{\blue{Illustrating more details of our approach. The token clustering layer consists of an adaptive average pooling block (for initializing the cluster centers) and an iterative local clustering block (for performing the k-means clustering). The token reconstruction layer consists of a token similarity estimation block (for estimating the reconstruction relation matrix) and a reconstruction block (for reconstructing the high-resolution representations). $\mathbf{Z}_\alpha$ represents the original high-resolution representations after $\alpha$-th transformer layer. $\mathbf{S}_\alpha$ represents the clustered low-resolution representations by token clustering layer. $\mathbf{S}_{\alpha+\beta}$ represents the refined clustered low-resolution representations after additional $\beta$ transformer layers. $\mathbf{Z}_{\alpha+\beta}$ represents the reconstructed high-resolution representations from $\mathbf{S}_{\alpha+\beta}$ by using the token reconstruction layer.}}}
\label{fig:approach_details}
\end{figure*}

\begin{table*}[t]
\begin{minipage}[t]{1\linewidth}
\centering
\footnotesize
\caption{{Illustrating the hyper-parameter settings used for Segmenter, DPT, and SWAG.}}
\resizebox{\linewidth}{!}{
\setlength\tabcolsep{3.2pt}
\renewcommand{\arraystretch}{1.5}
\begin{tabular}{l|l|l|c|c|c|c|c|c|c|c|c|c}
\shline
Method & Backbone & Dataset &  $\alpha$ & $\alpha+\beta$ & $\gamma$ & $\mathsf{L}$ & $\frac{\mathsf{H}}{\mathsf{P}}\times \frac{\mathsf{W}}{\mathsf{P}}$ &  $\mathsf{h}\times \mathsf{w}$ & $\lambda$ & $\kappa$ & $\tau$ & $\mathsf{k}$ \\
\shline
\multirow{3}{*}{Segmenter~\cite{Segmenter}} & \multirow{3}{*}{ViT-L/$16$} & ADE$20$K & $10$ & \multirow{3}{*}{$24$} & \multirow{3}{*}{$0$}  & \multirow{3}{*}{$24$} & $40\times40$ & $28\times28$ & \multirow{3}{*}{$5\times 5$} & \multirow{3}{*}{$5$} & \multirow{3}{*}{$50$} & \multirow{3}{*}{$20$} \\
& & Cityscapes & $12$ & & & & $48\times48$ & $32\times32$ & & & & \\
& & PASCAL-Context & $14$ & & & & $30\times30$ & $15\times15$ & & & & \\
\shline
\multirow{2}{*}{DPT~\cite{ranftl2021vision}} & \multirow{2}{*}{R$50$+ViT-B/$16$} & KITTI & $2$ & \multirow{2}{*}{$12$} & \multirow{2}{*}{$0$} & \multirow{2}{*}{$12$} & $76\times22$ & $28\times28$ & $5\times 5$ & $5$ & $5$ & $20$ \\
& & NYUv$2$ & $3$ &  & & & $40\times30$ & $16\times16$ & $7\times 7$ & $5$ & $10$ & $50$ \\
\shline
\multirow{2}{*}{SWAG~\cite{singh2022revisiting}} & ViT-H/$14$ & \multirow{2}{*}{ImageNet-$1$K} & $8$ & $32$ & $0$ & $32$ & $37\times37$ & $25\times25$ & $7\times 7$ & $5$ & $1$ & $20$ \\
& ViT-L/$16$ & & $8$ & $24$ & $0$ & $24$ & $32\times32$ & $22\times22$ & $9\times 9$ & $5$ & $1$ & $20$ \\
\shline
\end{tabular}
}
\label{tab:hyper-parameter-vit}
\vspace{1mm}
\end{minipage}
\begin{minipage}[t]{1\linewidth}
\centering
\renewcommand{\arraystretch}{1.8}
\footnotesize
\caption{{Illustrating the hyper-parameter settings used for Mask$2$Former and SwinV$2$-L + HTC++.}}
\resizebox{\linewidth}{!}{
\setlength\tabcolsep{1pt}
\begin{tabular}{l|l|l|c|c|c|c|c|c|c|c|c|c}
\shline
Method & Backbone & Dataset &  $\alpha$ & $\alpha+\beta$ & $\gamma$ & $\mathsf{L}$ & $\mathsf{K}\times \mathsf{K}$ &  $\mathsf{k}\times \mathsf{k}$ & $\lambda$ & $\kappa$ & $\tau$ & $\mathsf{k}$ \\
\shline
\multirow{3}{*}{Mask$2$Former~\cite{cheng2021mask2former}} & \multirow{3}{*}{Swin-L} & COCO (panoptic seg.) & $10$ & \multirow{3}{*}{$22$} & \multirow{3}{*}{$2$} & \multirow{3}{*}{$24$} & \multirow{3}{*}{$12\times 12$} & \multirow{3}{*}{$8\times 8$} & $7\times 7$ & $5$ & $20$ & $10$ \\ 
& & ADE20K (semantic seg.) & $8$ & & & & & & $5\times 5$ & $5$ & $100$ & $10$ \\
& & COCO (instance seg.) & $12$ & & & & & & $11\times 11$ & $5$ & $100$ & $60$ \\
\shline
SwinV$2$-L + HTC++~\cite{liu2021swinv2} & SwinV$2$-L & COCO (object det.) & $12$ & $22$ & $2$ & $24$ & $32\times32$ & $23\times23$ & $5\times 5$ & $5$ & $33$ & $20$ \\
\shline
\end{tabular}
}
\label{tab:hyper-parameter-swin}
\vspace{1mm}
\end{minipage}
\begin{minipage}[t]{1\linewidth}
\centering
\footnotesize
\caption{{Illustrating the hyper-parameter settings used for measuring FPS and GFLOPs.}}
\resizebox{\linewidth}{!}{
\setlength\tabcolsep{9pt}
\renewcommand{\arraystretch}{1.5}
\begin{tabular}{l|l|c|l|c}
\shline
Method & Backbone & with Head & Dataset &  Input resolution  \\
\shline
\multirow{3}{*}{Segmenter~\cite{Segmenter}} & \multirow{3}{*}{ViT-L/$16$} & \multirow{3}{*}{\checkmark} & ADE$20$K & $640\times640$\\
& & & Cityscapes & $768\times768$ \\
& & & PASCAL-Context & $480\times480$ \\
\shline
\multirow{2}{*}{DPT~\cite{ranftl2021vision}} & \multirow{2}{*}{R$50$+ViT-B/$16$} & \multirow{2}{*}{\checkmark} & KITTI & $1216\times352$ \\
& & & NYUv$2$ & $640\times480$ \\
\shline
\multirow{2}{*}{SWAG~\cite{singh2022revisiting}} & ViT-H/$14$ & \multirow{2}{*}{\checkmark} & \multirow{2}{*}{ImageNet-1K} & $518\times518$ \\
& ViT-L/$16$ & & & $512\times512$ \\
\shline
\multirow{3}{*}{Mask$2$Former~\cite{cheng2021mask2former}} & \multirow{3}{*}{Swin-L} & \multirow{3}{*}{\xmark} & COCO (panoptic seg.) & $1152\times1152$ \\ 
& & & ADE20K (semantic seg.) & $1152\times1152$ \\
& & & COCO (instance seg.) & $1152\times1152$ \\
\shline
SwinV$2$-L + HTC++~\cite{liu2021swinv2} & SwinV$2$-L & \xmark & COCO (object det.) & $1024\times1024$ \\
\shline
\end{tabular}
}
\label{tab:eval_details}
\end{minipage}
\begin{minipage}[t]{1\linewidth}
\centering
\setlength\tabcolsep{10pt}
\footnotesize
\caption{Comparison to parametric methods based on Segmenter~\cite{Segmenter}.}
\begin{tabular}{l|l|cc|cc}
\shline
Dataset & Method  & Parametric & Fine-Tuning & GFLOPs & mIoU \\
\hline
\multirow{6}{*}{ADE$20$K} & {Dynamic ViT ($\rho=0.7$)} & \checkmark & \checkmark & $455.6$ & $45.62$ \\
& {Dynamic ViT ($\rho=0.8$)} & \checkmark & \checkmark & $513.3$ & $47.89$ \\
& {Dynamic ViT ($\rho=0.9$)} & \checkmark & \checkmark & $583.0$ & $50.42$ \\\cline{2-6}
& {Ours ($\mathrm{h}\times\mathrm{w}=16\times16$)} & \xmark & \xmark & $315.1$ & $48.21$ \\
& {Ours ($\mathrm{h}\times\mathrm{w}=20\times20$)} & \xmark & \xmark & $347.2$ & $50.17$ \\
& {Ours ($\mathrm{h}\times\mathrm{w}=24\times24$)} & \xmark & \xmark & $388.2$ & $51.32$ \\
\shline
\end{tabular}
\label{table:somparison_with_dynamic_vit}
\end{minipage}
\begin{minipage}[t]{0.485\linewidth}
\centering
\footnotesize
\setlength{\tabcolsep}{1pt}
\renewcommand{\arraystretch}{1}
\caption{\small{Illustrating the differences between clustered attention~\cite{vyas2020fast}, ACT~\cite{zheng2020end}, SMRF~\cite{daras2020smyrf}, and our approach.}}
\resizebox{\linewidth}{!}
{
\begin{tabular}{ l | c | c | c | c}
\shline
\pbox{22mm}{Cluster method}   & query   &  key-value & FFN & \#clustering layers  \\ \hline
Clustered Attention~\cite{vyas2020fast}  &  \checkmark  &   \xmark & \xmark & \# MHSA layers   \\
ACT~\cite{zheng2020end} &  \checkmark  &   \xmark & \xmark & \# MHSA layers   \\
SMRF~\cite{daras2020smyrf} &  \checkmark  &   \checkmark & \xmark & \# MHSA layers   \\
Ours &  \checkmark  &   \checkmark & \checkmark & $1$   \\
\shline
\end{tabular}}
\label{tab:compare_clusterattn_act_smrf}
\end{minipage}
\begin{minipage}[t]{0.485\linewidth}
\centering
\footnotesize
\caption{\small{Comparison results with ACT~\cite{zheng2020end}.}}
\setlength{\tabcolsep}{0.4pt}
\renewcommand{\arraystretch}{1.1}
\resizebox{\linewidth}{!}
{
\begin{tabular}{ l | c | c | c }
\shline
\pbox{22mm}{Cluster method}   &  FPS   &  GFLOPs & mIoU \\ \hline
Segmenter+ViT-B/$16$           &  $6.2$ & $659.0$ &  $51.82$   \\
Segmenter+ViT-B/$16$+Ours($\mathrm{h}$$\times$$\mathrm{w}$=$24\times24$)  &  $9.1$ & $388.2$ & $51.32$  \\
Segmenter+ViT-B/$16$+Ours($\mathrm{h}$$\times$$\mathrm{w}$=$28\times28$)  &  $8.8$ & $438.9$ & $51.56$  \\
Segmenter+ViT-B/$16$+ACT(\#query-hashes=$16$)  &  $5.8$ & $578.7$ & $48.12$  \\
Segmenter+ViT-B/$16$+ACT(\#query-hashes=$24$)  &  $5.3$ & $614.7$ & $51.38$  \\
Segmenter+ViT-B/$16$+ACT(\#query-hashes=$32$)  &  $5.0$ & $638.2$ & $51.64$  \\
\shline
\end{tabular}}
\label{tab:compare_act}
\end{minipage}
\vspace{-5mm}
\end{table*}

\section*{A. Illustrating More Details of Our Approach}
\blue{We first illustrate the overall details of our token clustering layer and token reconstruction layer in Figure~\ref{fig:approach_details}.
We then present the example implementation of token clustering layer and token reconstruction layer based on PyTorch in Listing \ref{lst:token_clustering} and Listing \ref{lst:token_reconstruct}, respectively.}

\section*{B. More Hyper-parameter Details}
We summarize the detailed hyper-parameter settings for the dense prediction methods based on plain ViTs and Swin Transformers in Table~\ref{tab:hyper-parameter-vit} and Table~\ref{tab:hyper-parameter-swin}, respectively.

Table~\ref{tab:hyper-parameter-vit} summarizes the hyper-parameters,
including the inserted positions $\alpha$ \& $\alpha$+$\beta$ of token clustering layer \& token reconstruction layer, the number of remaining transformer layers after the token reconstruction layer $\gamma$, the total number of transformer layers $\mathsf{L}$, the number of tokens before clustering $\frac{\mathsf{H}}{\mathsf{P}}\times \frac{\mathsf{W}}{\mathsf{P}}$, the number of tokens after clustering $\mathsf{h}\times \mathsf{w}$, the number of neighboring pixels $\lambda$, the number of EM iterations $\kappa$, the temperature value $\tau$, and the number of nearest neighbors $\mathsf{k}$,
for Segmenter, DPT, and SWAG.

Table~\ref{tab:hyper-parameter-swin} summarizes the hyper-parameters, including the inserted positions $\alpha$ \& $\alpha$+$\beta$ of the window token clustering layer \& window token reconstruction layer, the number of remaining transformer layers after the token reconstruction layer $\gamma$, the total number of transformer layers $\mathsf{L}$, the number of window tokens before clustering $\mathsf{K}\times \mathsf{K}$, the number of window tokens after clustering $\mathsf{k}\times \mathsf{k}$, the number of neighboring pixels $\lambda$, the number of EM iterations $\kappa$, the temperature value $\tau$, and the number of nearest neighbors $\mathsf{k}$, for Mask$2$Former and SwinV$2$-L + HTC++.

\section*{C. More Evaluation Details}
We illustrate the evaluation details used for measuring the
GFLOPs and FPS of different methods in Table~\ref{tab:eval_details}.
We choose the input resolutions for different methods with different backbones according to their official implementations.
To illustrate the effectiveness of our method more accurately, we do not include the complexity and latency
brought by the especially heavy detection heads or segmentation heads
within Mask$2$Former and SwinV$2$-L + HTC++.
For example, the GFLOPs of SwinV$2$-L backbone accounts for only $56.7\%$
of the whole model, therefore, we only report the GFLOPs and FPS improvements of our method over the backbone.

\section*{D. Comparison with EViT~\cite{liang2021evit} on Dense Prediction}
\begin{figure*}[h]
\vspace{-3mm}
\begin{minipage}[t]{\linewidth}
\centering
\begin{subfigure}[b]{1\textwidth}
\centering
\caption{\small{Adapting EViT for Dense Prediction}}
\hspace{-10mm}
{\includegraphics[width=\textwidth]{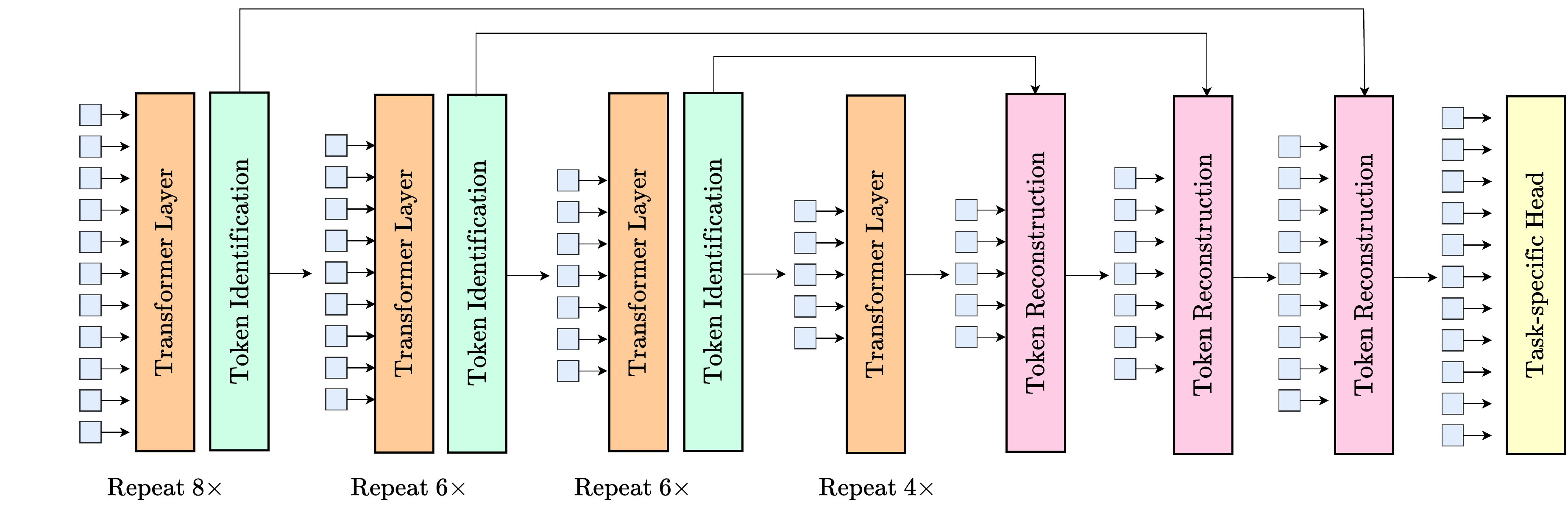}}
\vspace{1mm}
\label{fig:dense_evit}
\end{subfigure}
\begin{subfigure}[b]{1\textwidth}
\centering
\caption{\small{Adapting DynamicViT for Dense Prediction}}
{\includegraphics[width=0.945\textwidth]{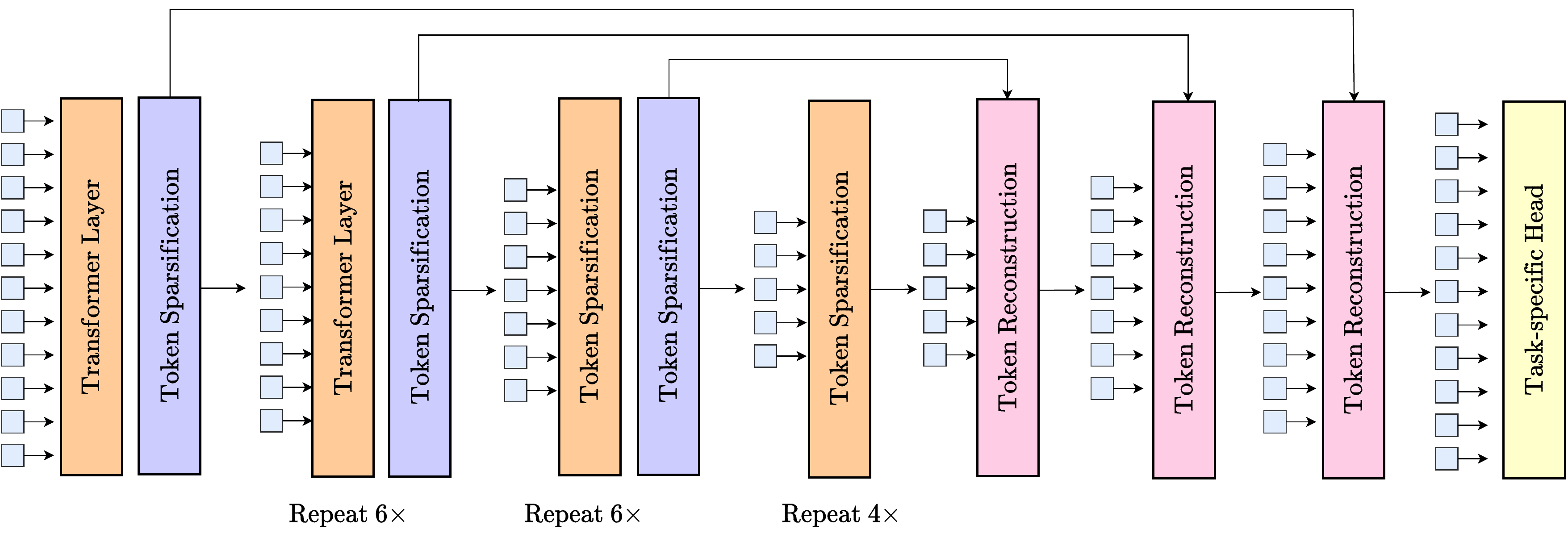}}
\end{subfigure}
\caption{{\small Illustrating how to adapt the EViT~\cite{liang2021evit} and DynamicViT~\cite{rao2021dynamicvit} for dense prediction based on ViT-L/$16$ with $24$ transformer layers. Following the proposed token reconstruction scheme, we estimate the semantic relations based on the representations before each token identification layer~\cite{liang2021evit} or token sparsification layer~\cite{rao2021dynamicvit}}.}
\label{fig:dense_dyvit}
\end{minipage}
\begin{minipage}[t]{\linewidth}
\centering
\begin{subfigure}[b]{0.4\textwidth}
\centering
{\includegraphics[width=\textwidth]{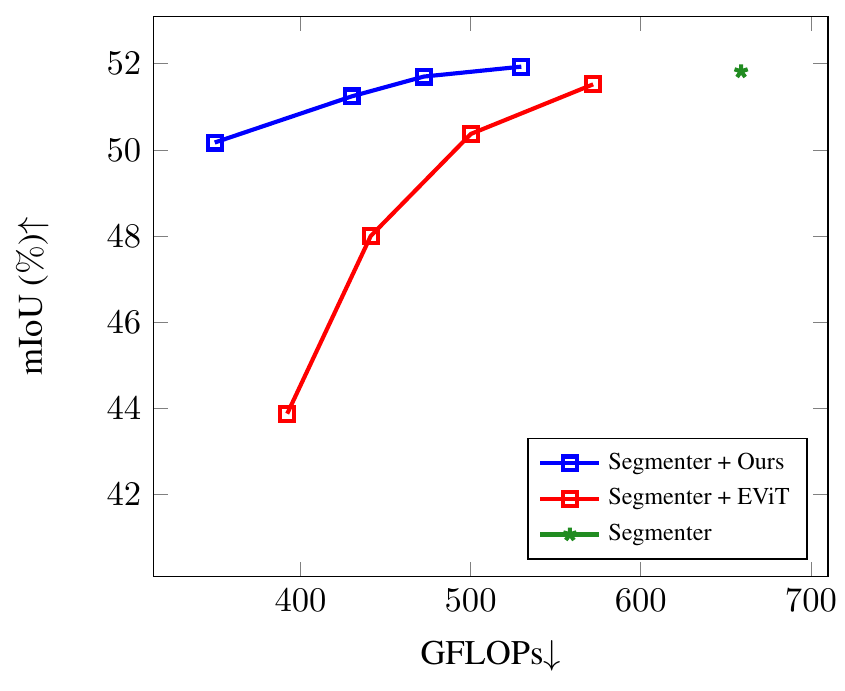}}
\caption{\small{mIoU vs. GFLOPs on ADE$20$K}}
\end{subfigure}
\hspace{10mm}
\begin{subfigure}[b]{0.4\textwidth}
\centering
{\includegraphics[width=\textwidth]{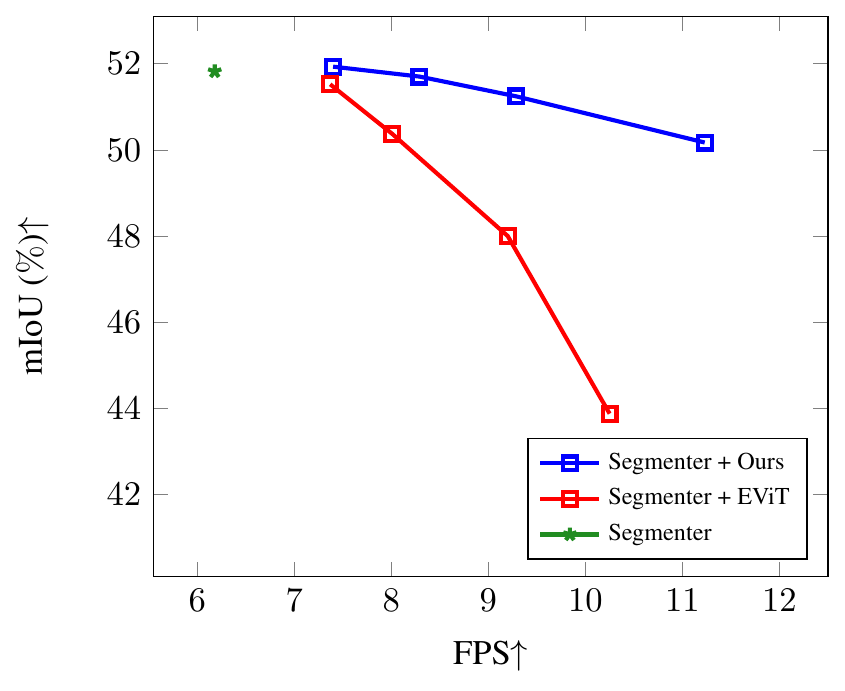}}
\caption{\small{mIoU vs. FPS on ADE$20$K}}
\end{subfigure}
\label{fig:dense_evit_dyvit}
\end{minipage}
\\
\begin{minipage}[t]{\linewidth}
\centering
\begin{subfigure}[b]{0.4\textwidth}
\centering
{\includegraphics[width=\textwidth]{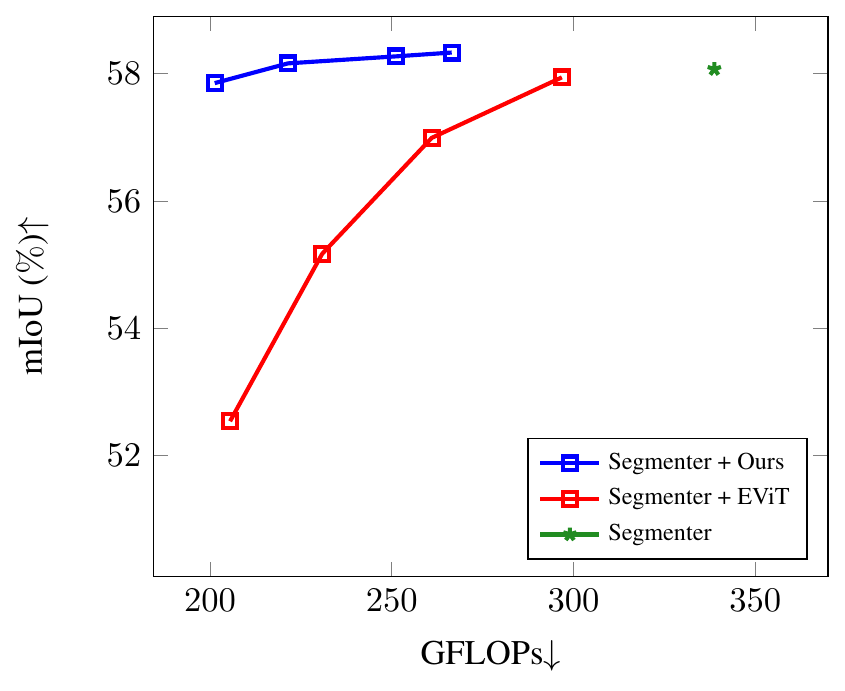}}
\caption{\small{mIoU vs. GFLOPs on PASCAL-Context}}
\end{subfigure}
\hspace{10mm}
\begin{subfigure}[b]{0.4\textwidth}
\centering
{\includegraphics[width=\textwidth]{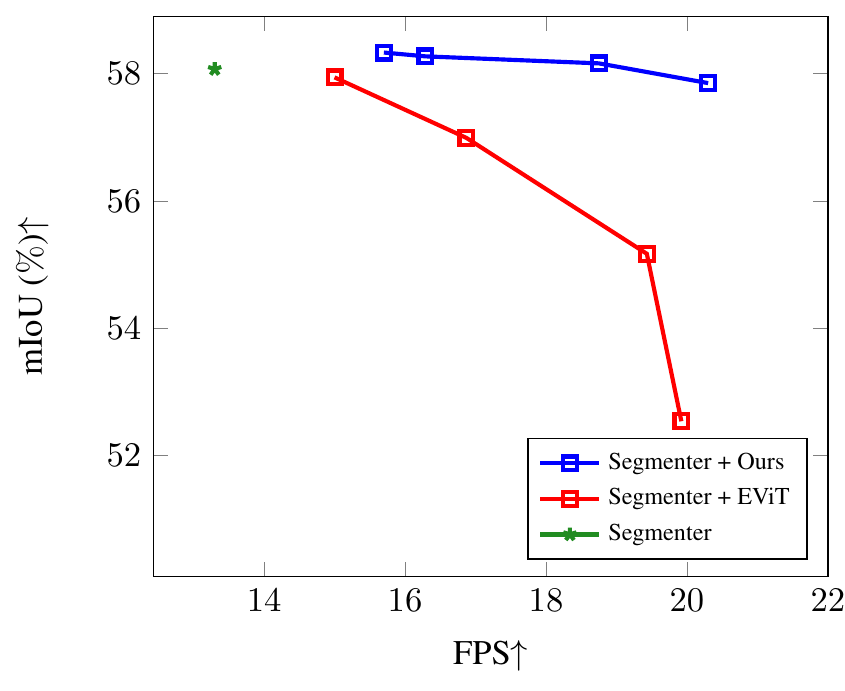}}
\caption{\small{mIoU vs. FPS on PASCAL-Context}}
\end{subfigure}
\label{fig:compare_gflops}
\end{minipage}
\caption{\small{Comparison with EViT~\cite{liang2021evit} on ADE$20$K and PASCAL-Context semantic segmentation task based on Segmenter with ViT-L/$16$. $\uparrow$ and $\downarrow$ represent higher is better and lower is better respectively.}}
\label{fig:compare_evit}
\end{figure*}

To demonstrate the advantage of our approach over the representative method that is originally designed for the image classification tasks, i.e., EViT~\cite{liang2021evit},
we report the detailed comparison results in Figure~\ref{fig:compare_evit}.
The original EViT propose to identify and only keep the top $\rho\%$ tokens according to their attention scores relative to the [\texttt{class}] token.
Specifically, we follow the official implementations to insert the token identification module into the $8$-th, $14$-th, and $20$-th layer
of ViT-L/$16$ (with $24$ layers in total) to decrease the number of tokens by ($1$-$\rho\%$), respectively.
We report the results of EViT by choosing $\rho\%$=$60\%$/$70\%$/$80\%$/$90\%$ in Figure~\ref{fig:compare_evit}.
Accordingly, we can see that our method significantly outperforms EViT across various GFLOPs \& FPS settings when evaluating without either re-training or fine-tuning.

The EViT can not be used for dense prediction directly, as it only keeps around $21.6\%\sim72.9\%$ of the tokens at last.
To reconstruct the missed token representations over the abandoned positions,
we apply two different strategies, including (i) reusing the representations before the corresponding token identification module, and (ii) using our token reconstruction layer to reconstruct the missed token representations according to Figure~\ref{fig:dense_evit}.
We empirically find the first strategy achieves much worse results, thus choosing the second strategy by default.

\section*{E. Adapting DynamicViT~\cite{rao2021dynamicvit} for Dense Prediction}

To adapt DynamicViT for dense prediction tasks, we propose to add multiple token reconstruction layers to reconstruct high-resolution representations from the selected low-resolution representations iteratively.
Figure~\ref{fig:dense_dyvit} (b) presents more details of the overall framework.
We also report the comparison results in Table~\ref{table:somparison_with_dynamic_vit}.

\section*{F. Comparison with Clustered Attention~\cite{vyas2020fast}, ACT~\cite{zheng2020end}, and SMRF~\cite{daras2020smyrf}}

We illustrate the key differences between our approach and the existing clustered attention approaches~\cite{vyas2020fast,zheng2020end,daras2020smyrf} the following two aspects:
(i) These clustering attention methods perform clustering within each multi-head self-attention layer (MHSA) independently while our approach only performs clustering once with the token clustering layer and refines the clustered representations with the following transformer layers. Therefore, our approach introduces a much smaller additional overhead caused by the clustering operation.
(ii) These clustering attention methods only reduce the computation cost of each MHSA layer equipped with clustering attention as they maintain the high-resolution representations outside the MHSA layers while Our approach can reduce the computation cost of both MHSA layers and feed-forward network (FFN) layers after the token clustering layer.
We further summarize their detailed differences and the experimental comparison resultswith ACT~\cite{zheng2020end}(without retraining) in Table~\ref{tab:compare_clusterattn_act_smrf} and Table~\ref{tab:compare_act}, respectively.

According to the results in Table~\ref{tab:compare_act}, we can see that (i) ACT also achieves strong performance without retraining, (ii) our approach is a better choice considering the trade-off between performance and FPS \& GFLOPs, e.g., our method achieves close performance as ACT ($51.32$ vs. $51.38$) while running $70\%$ faster ($9.1$ vs. $5.3$) and saving more than $35\%$ GFLOPs ($388.2$ vs. $614.7$).

\section*{G. Visualization}
\begin{figure*}[t]
\begin{minipage}[t]{\linewidth}
\centering
\begin{subfigure}[b]{1\textwidth}
\centering
\includegraphics[width = 0.2\textwidth]{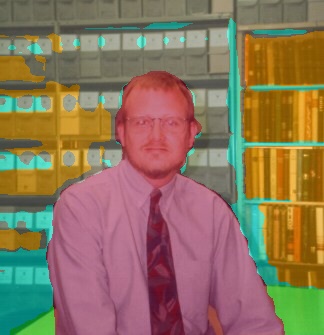}
\includegraphics[width = 0.2\textwidth]{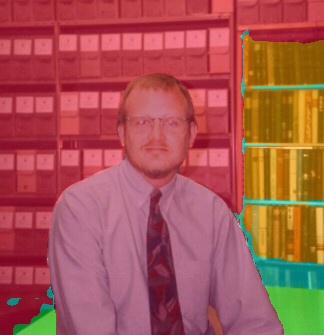}
\includegraphics[width = 0.2\textwidth]{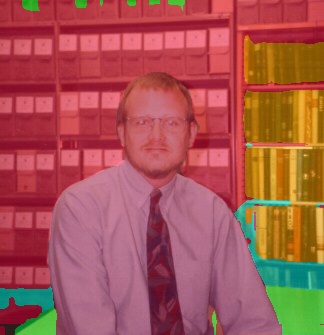}
\includegraphics[width = 0.2\textwidth]{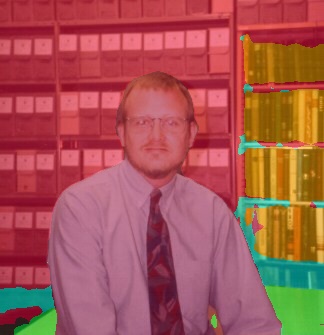}\\\vspace{1mm}
\includegraphics[width = 0.2\textwidth]{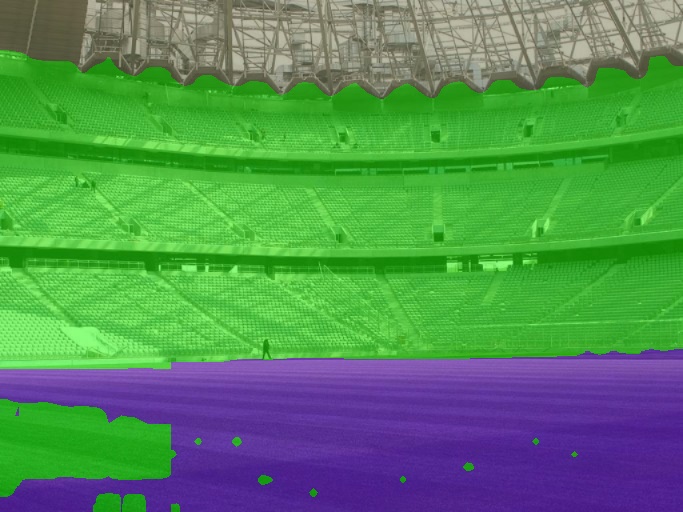}
\includegraphics[width = 0.2\textwidth]{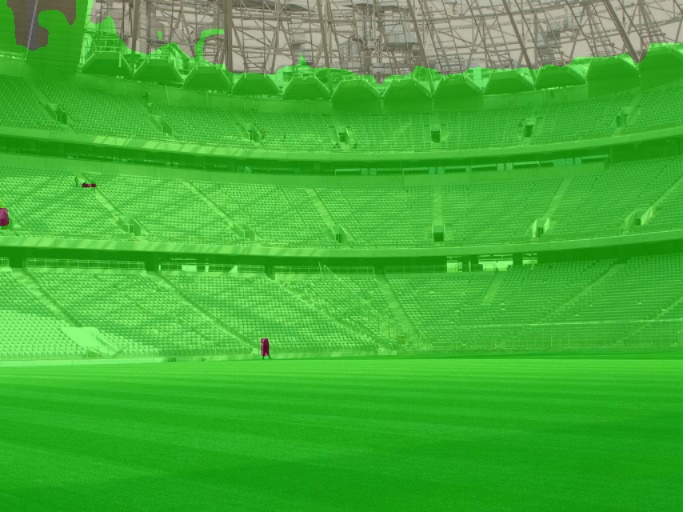}
\includegraphics[width = 0.2\textwidth]{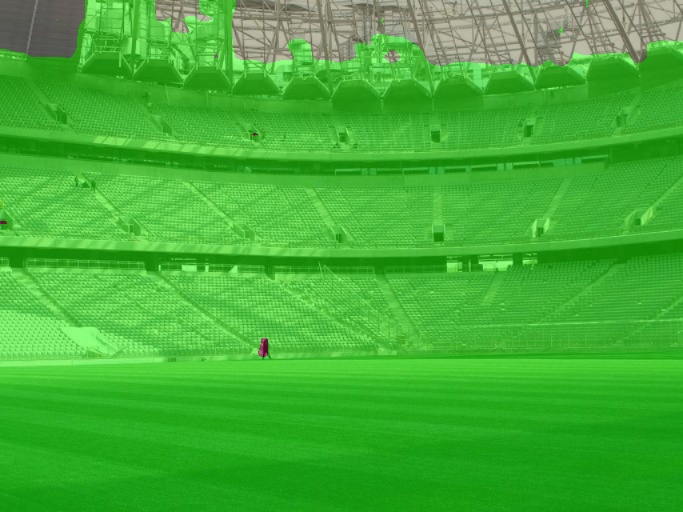}
\includegraphics[width = 0.2\textwidth]{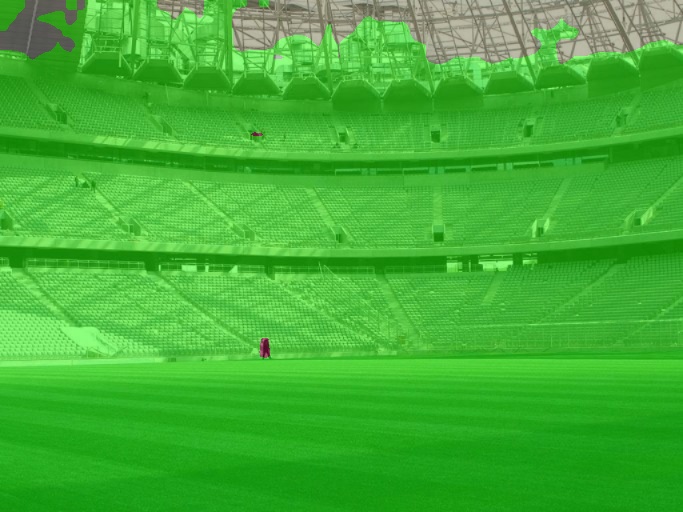}\\\vspace{1mm}
\includegraphics[width = 0.2\textwidth]{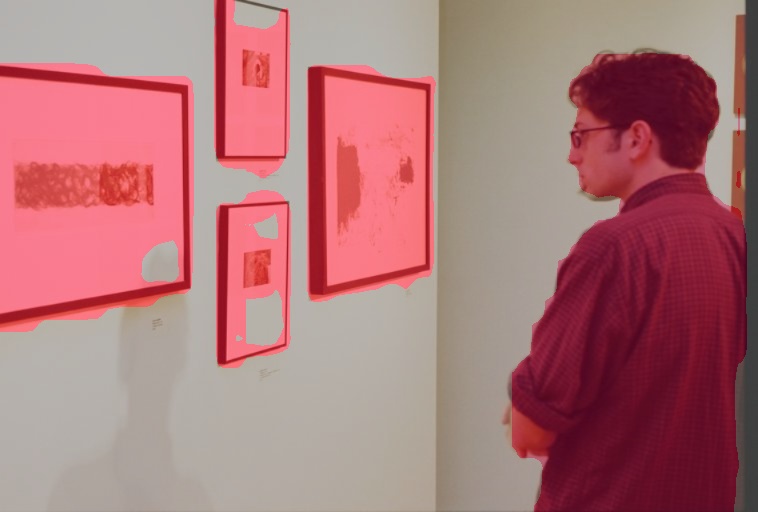}
\includegraphics[width = 0.2\textwidth]{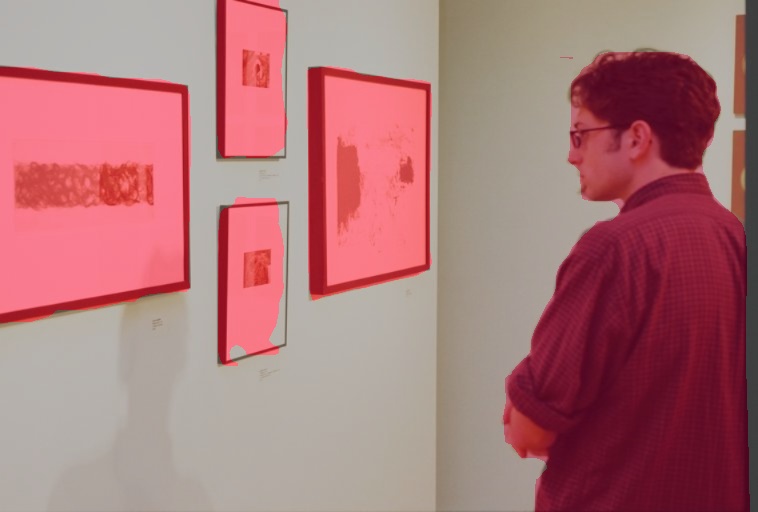}
\includegraphics[width = 0.2\textwidth]{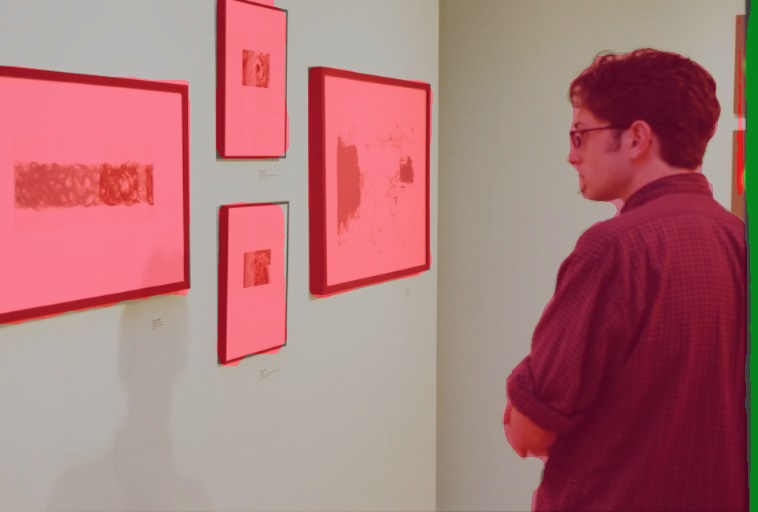}
\includegraphics[width = 0.2\textwidth]{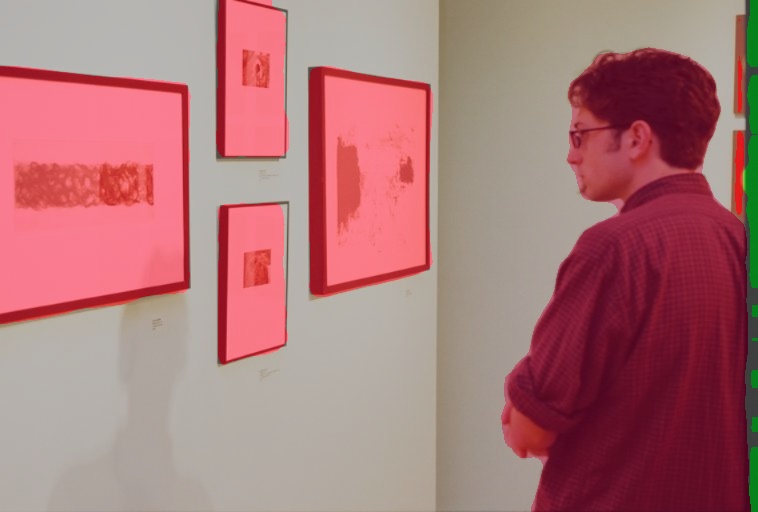}
\caption{\small{ADE$20$K example segmentation results of our approach with $\rm{h}\times \rm{w}$ as $8\times8$, $16\times16$, $24\times24$ on the left three columns, respectively. The right-most column shows the results of the original Segmenter+ViT-L/$16$. We can see that our approach achieves consistently better segmentation results with increasing clustered output resolutions from left to right.}}
\label{fig:vis_seg_results}
\end{subfigure}
\vspace{2mm}
\end{minipage}
\begin{minipage}[t]{\linewidth}
\centering
\begin{subfigure}[b]{1\textwidth}
\centering
\includegraphics[width = 0.2\textwidth,height=20mm]{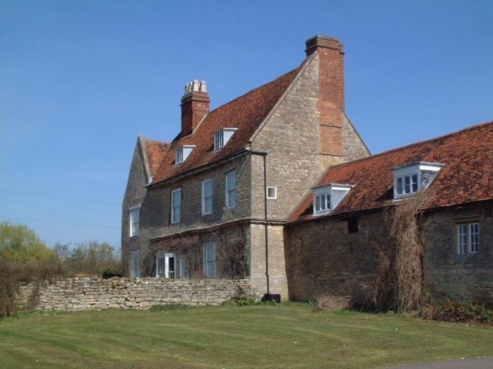}
\includegraphics[width = 0.2\textwidth,height=20mm]{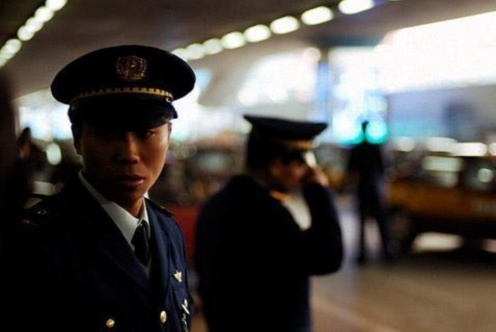}
\includegraphics[width = 0.2\textwidth,height=20mm]{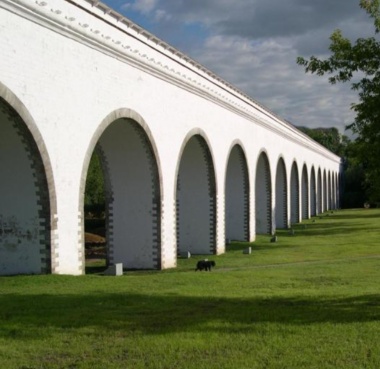}
\includegraphics[width = 0.2\textwidth,height=20mm]{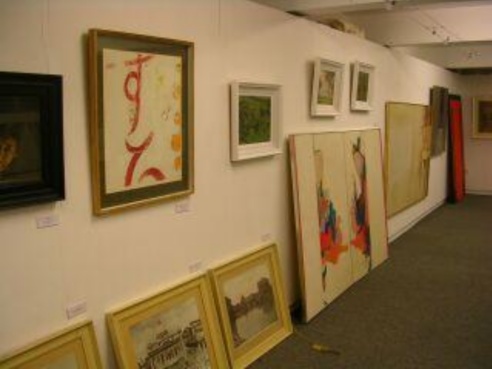}\\\vspace{1mm}
\includegraphics[width = 0.2\textwidth,height=20mm]{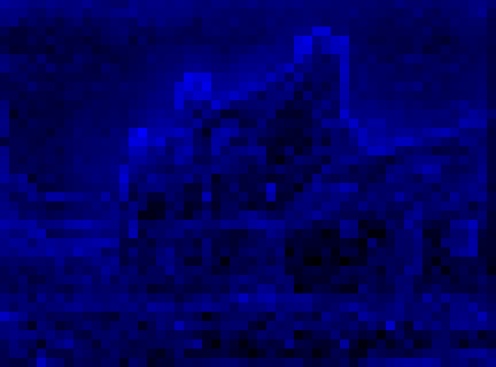}
\includegraphics[width = 0.2\textwidth,height=20mm]{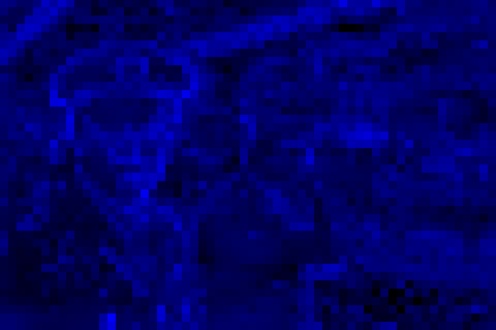}
\includegraphics[width = 0.2\textwidth,height=20mm]{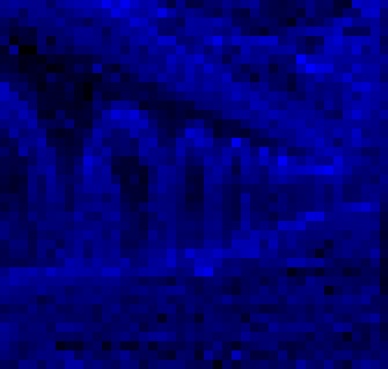}
\includegraphics[width = 0.2\textwidth,height=20mm]{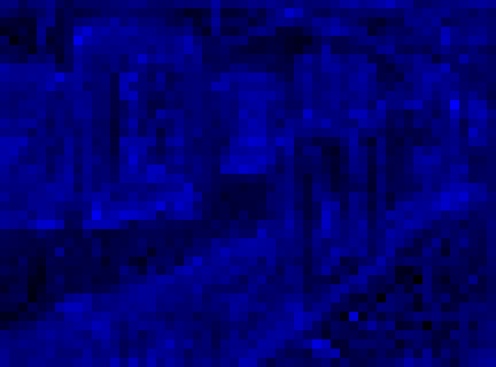}\\\vspace{1mm}
\includegraphics[width = 0.2\textwidth,height=25mm]{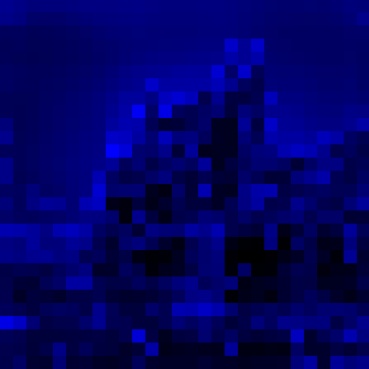}
\includegraphics[width = 0.2\textwidth,height=25mm]{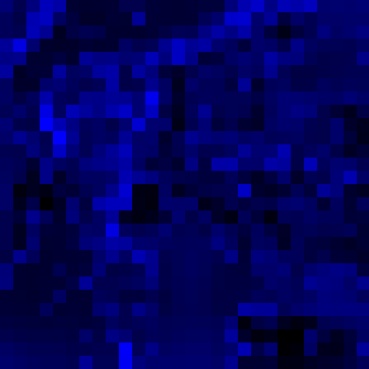}
\includegraphics[width = 0.2\textwidth,height=25mm]{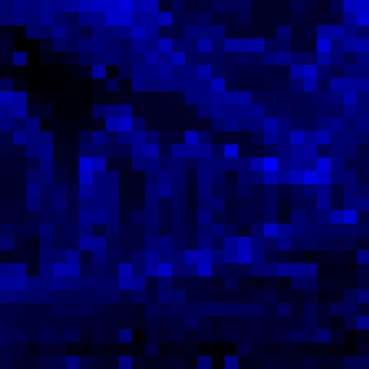}
\includegraphics[width = 0.2\textwidth,height=25mm]{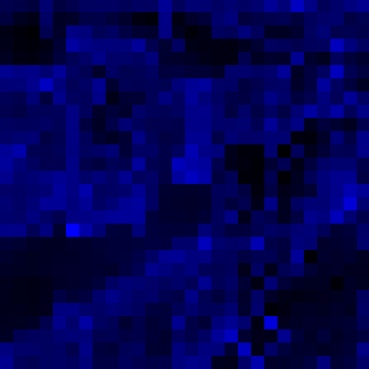}
\caption{\small{ADE$20$K example visualization of the original feature maps ($2$-ed row) and the clustering feature maps ($3$-rd row). We can see that the clustering feature maps still maintain the structure information presented in the original high-resolution feature maps, thus showing the potential benefits of our token clustering scheme.}}
\label{fig:vis_feat}
\end{subfigure}
\end{minipage}
\caption{\small{Visualizations of segmentation results in (a) and feature maps in (b). We choose Segmenter+ViT-L/$16$ on ADE$20$K to generate the above segmentation results and the feature map visualizations.}}
\label{fig:vis_results}
\end{figure*}

\blue{
We first present the visual comparison results of our approach in Figure~\ref{fig:vis_seg_results},
which shows three different configurations over Segmenter+ViT-L/$16$ achieve $32.13\%$/$48.21\%$/$51.32\%$ when setting the cluster size $\rm{h}\times \rm{w}$ as $8\times8$/$16\times16$/$24\times24$, respectively.}

\blue{
Then, we visualize both the original feature maps and the clustering feature maps in Figure~\ref{fig:vis_feat}.
Accordingly, we can see that the clustering feature maps, based on our token clustering layer, well maintain the overall structure information carried in the original high-resolution feature maps.
}

\begin{figure*}[t]
\begin{minipage}[t]{\linewidth}
\centering
\begin{subfigure}[b]{1\textwidth}
\centering
\includegraphics[width = 0.2\textwidth,height=20mm]{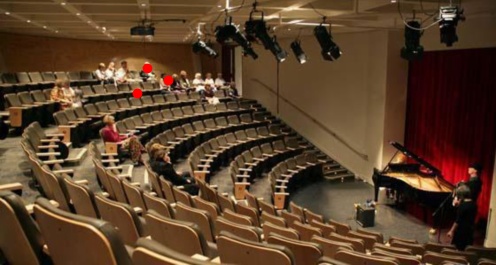}
\includegraphics[width = 0.2\textwidth,height=20mm]{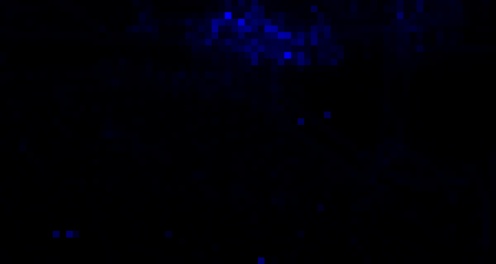}
\includegraphics[width = 0.2\textwidth,height=20mm]{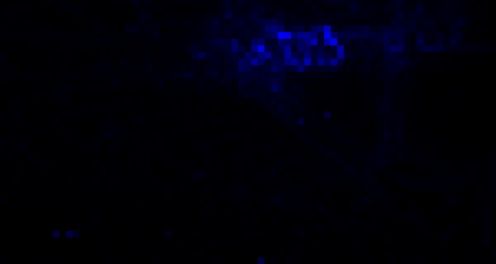}
\includegraphics[width = 0.2\textwidth,height=20mm]{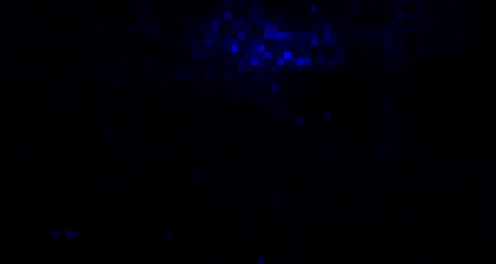}\\\vspace{1mm}
\includegraphics[width = 0.2\textwidth,height=20mm]{img/vis_attention/img2/sample_point.jpg}
\includegraphics[width = 0.2\textwidth,height=20mm]{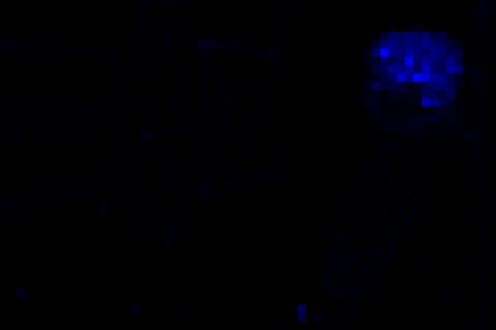}
\includegraphics[width = 0.2\textwidth,height=20mm]{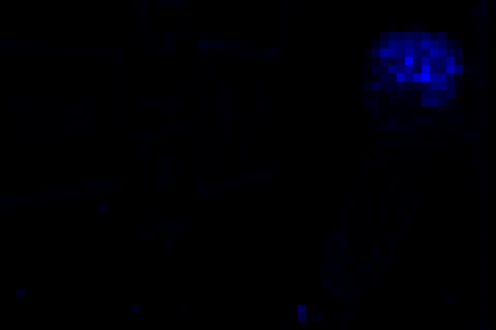}
\includegraphics[width = 0.2\textwidth,height=20mm]{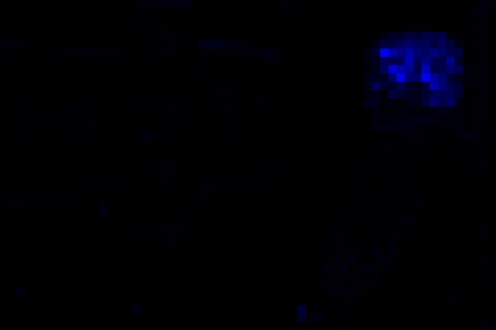}\\\vspace{1mm}
\includegraphics[width = 0.2\textwidth,height=20mm]{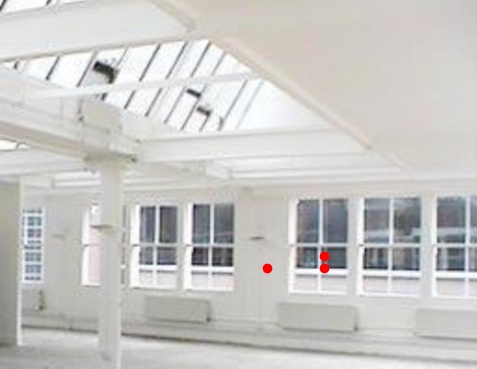}
\includegraphics[width = 0.2\textwidth,height=20mm]{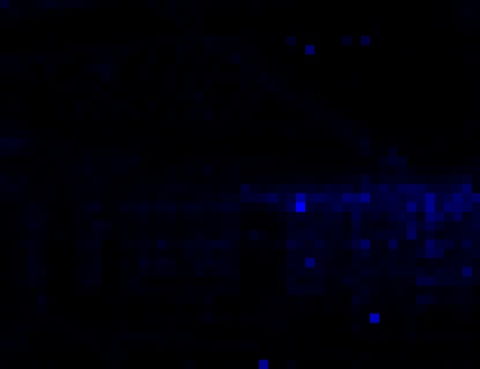}
\includegraphics[width = 0.2\textwidth,height=20mm]{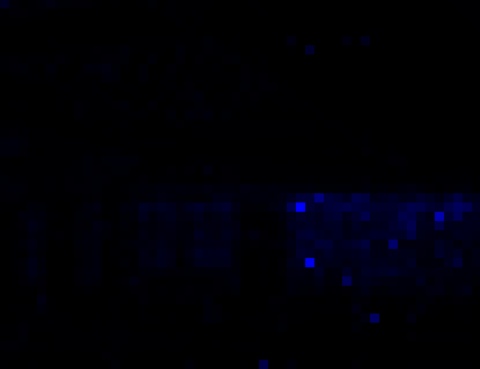}
\includegraphics[width = 0.2\textwidth,height=20mm]{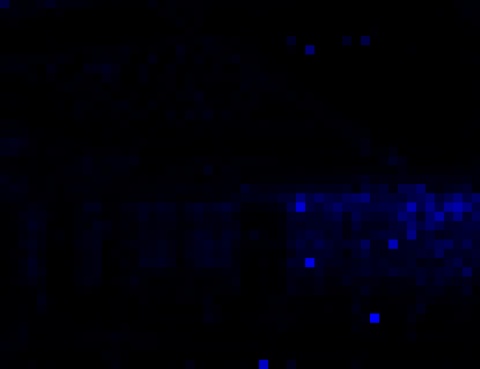}\\\vspace{1mm}
\includegraphics[width = 0.2\textwidth,height=20mm]{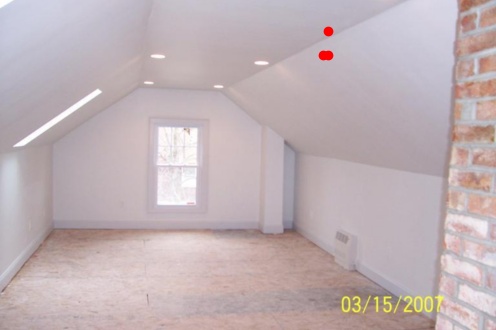}
\includegraphics[width = 0.2\textwidth,height=20mm]{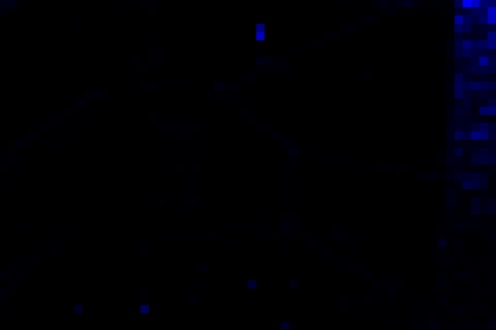}
\includegraphics[width = 0.2\textwidth,height=20mm]{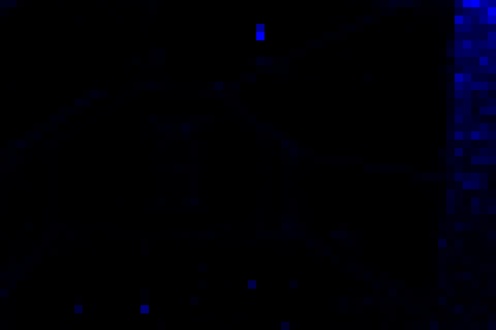}
\includegraphics[width = 0.2\textwidth,height=20mm]{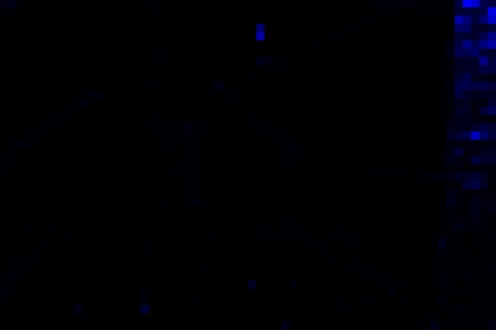}\\\vspace{1mm}
\includegraphics[width = 0.2\textwidth,height=20mm]{img/vis_attention/img5/sample_point.jpg}
\includegraphics[width = 0.2\textwidth,height=20mm]{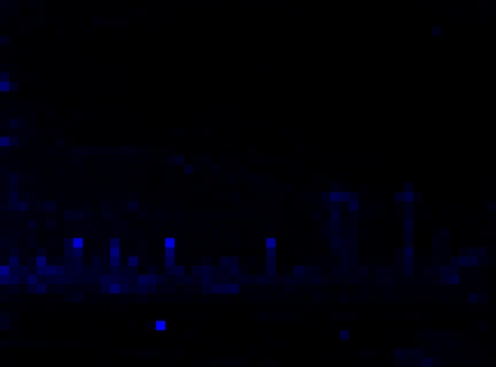}
\includegraphics[width = 0.2\textwidth,height=20mm]{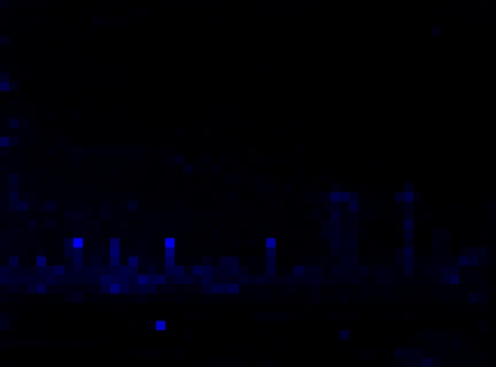}
\includegraphics[width = 0.2\textwidth,height=20mm]{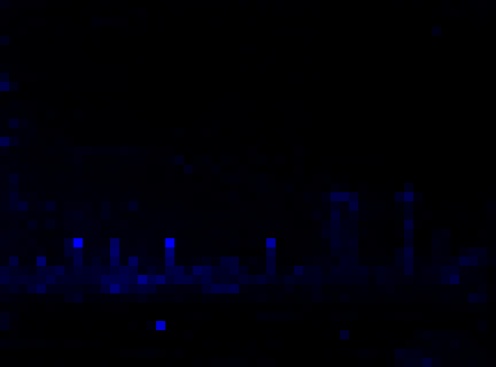}
\end{subfigure}
\end{minipage}
\caption{\small{Visualizations of the attention maps of neighboring sampled positions. We mark the sampled positions with red point markers. We can see that the neighboring positions share highly similar attention maps, which matches the redundancy observation in the Adaptive Clustering Transformer (ACT)~\cite{zheng2020end}.}}
\label{fig:vis_local_attention}
\end{figure*}

\blue{
Last, to verify the redundancy in the tokens of vision transformer, we visualize the attention maps of neighboring tokens in Figure~\ref{fig:vis_local_attention}.
}

\begin{minipage}{\linewidth}
\begin{lstlisting}[language=Python,label={lst:token_clustering},caption=\small{PyTorch example of token clustering layer.}]
def token_clustering_layer(features, cluster_features_shape, num_iters, tau):
    # args:
    #       features: shape [B, C, H, W]
    #       cluster_features_shape: [h, w]
    #       num_iters: num of iterations of updating cluster features
    #       teu: the temperture of distance matrix
    # output:
    #       cluster_features: shape [B, hw, C]
    
    B, C, H, W = features.shape
    
    # initialize the cluster features
    cluster_features = interpolate(features, cluster_features_shape)
    
    # construct mask to constrain the interactions within local range
    mask = calculate_mask(features.shape, cluster_features_shape)
    mask = (~mask) * 1e16

    features = features.reshape(B, C, -1).permute(0, 2, 1)  # (B, HW, C)
    cluster_features = cluster_features.reshape(B, C, -1).permute(0, 2, 1)  # (B, hw, C)

    for _ in range(num_iters):
        # calculate L2 distance of features and cluster features, the shape distance_matrix is (B, hw, HW)
        distance_matrix = L2_distance(features, cluster_features)
        # mask remote distance through softmax
        distance_matrix += mask
        weights = (-distance_matrix / teu).softmax(dim=1)
        # let the sum of weight of each cluster feature be 1
        weights = weights / weights.sum(dim=2, keepdim=True).clamp_(min=1e-16)
        cluster_features = matrix_product(weights, features)

    return cluster_features
\end{lstlisting}
\end{minipage}

\begin{minipage}{\linewidth}
\begin{lstlisting}[language=Python,label={lst:token_reconstruct},caption=\small{PyTorch example of token reconstruction layer.}]
def token_reconstruction_layer(cluster_features, features_before_clustering, features_after_clustering, k, teu):
    # args:
    #       cluster_features: shape [B, hw, C]
    #       features_before_clustering: features of alpha-th layer before clustering, shape [B, hw, C]
    #       features_after_clustering: features of alpha-th layer before clustering, shape [B, HW, C]
    #       k: topk parameter
    #       teu: the temperture of weight matrix
    # output:
    #       features: reconstruction features, shape [B, HW, C]

    # calculate L2 distance between features and cluster_features
    distance = L2_distance(features_before_clustering, features_after_clustering)  
    weight = exp(-teu * distance)
    # only remain the k weight of the most simliar features, calculating mask
    topk, indices = topk(weight, k=k, dim=2)
    mink = min(topk, dim=-1).values
    mink = mink.unsqueeze(-1).repeat(1, 1, weight.shape[-1])
    mask = greater_or_equal(weight, mink)
    weight = weight * mask
    
    weight = weight / weight.sum(dim=2, keepdim=True).clamp_(min=1e-16)
    features = matrix_product(weight, cluster_features)

    return features
\end{lstlisting}
\end{minipage}

\end{document}